\documentclass[letterpaper]{article} 
\usepackage{aaai24}  
\usepackage{times}  
\usepackage{helvet}  
\usepackage{courier}  
\usepackage[hyphens]{url}  
\usepackage{graphicx} 
\usepackage{natbib}  
\usepackage{caption} 
\usepackage{algorithm}
\usepackage{algorithmic}
\usepackage{amssymb}
\usepackage{amsmath}
\usepackage{multirow}
\usepackage{booktabs}
\usepackage{array}
\usepackage{makecell}
\usepackage{color}
\newcommand{\hut}{\textcolor{black}}
\newcommand{\yr}{\textcolor{black}}

\usepackage{xspace}

\def \pzo {\phantom{0}} 
\usepackage{newfloat}
\usepackage{listings}
\DeclareCaptionStyle{ruled}{labelfont=normalfont,labelsep=colon,strut=off} 
\floatstyle{ruled}
\newfloat{listing}{tb}{lst}{}
\floatname{listing}{Listing}
\nocopyright 

\title{AnomalyDiffusion: Few-Shot Anomaly Image Generation with Diffusion Model}

\author{Teng Hu$^{1}$\footnotemark[1],~~~Jiangning Zhang$^{2}$\thanks{Equal contributions.},~~~Ran Yi$^{1}$\thanks{Corresponding author.},~~~Yuzhen Du$^{1}$,~~~Xu Chen$^{2}$,\\~~~Liang Liu$^{2}$,~~~Yabiao Wang$^{2}$,~~~Chengjie Wang$^{1,2}$
\\ 
$^1$Shanghai Jiao Tong University~~~$^2$Youtu Lab, Tencent\\
\tt\small $\{$hu-teng, ranyi, Haaaaaaaaaa$\}$@sjtu.edu.cn;\\
\tt\small $\{$vtzhang,  cxxuchen, leoneliu, caseywang, jasoncjwang$\}$@tencent.com;\\
}

\usepackage{bibentry}

\begin{document}

\maketitle

\begin{abstract}
Anomaly inspection plays an important role in industrial manufacture. Existing anomaly inspection methods are limited in their performance due to insufficient anomaly data. 
Although anomaly generation methods have been proposed to augment the anomaly data,
they either suffer from poor generation authenticity or inaccurate alignment between the generated anomalies and masks. To address the above problems, we propose \textit{AnomalyDiffusion}, a novel diffusion-based few-shot anomaly generation model, which utilizes the strong prior information of latent diffusion model learned from large-scale dataset to enhance the generation authenticity under few-shot training data. Firstly, we propose Spatial Anomaly Embedding, which consists of a learnable anomaly embedding and a spatial embedding encoded from an anomaly mask, disentangling the anomaly information into anomaly appearance and location information. 
Moreover, to improve the alignment between the generated anomalies and the anomaly masks, we introduce a novel Adaptive Attention Re-weighting Mechanism. Based on the disparities between the generated anomaly image and normal sample, it dynamically guides the model to focus more on the areas with less noticeable generated anomalies, enabling generation of accurately-matched anomalous image-mask pairs. Extensive experiments demonstrate that our model significantly outperforms the state-of-the-art methods in generation authenticity and diversity, and effectively improves the performance of downstream anomaly inspection tasks. The code and data are available in
\url{https://github.com/sjtuplayer/anomalydiffusion}.



\end{abstract}

\section{Introduction}

\begin{figure}[t!]
\centering
\includegraphics[width=0.43\textwidth]{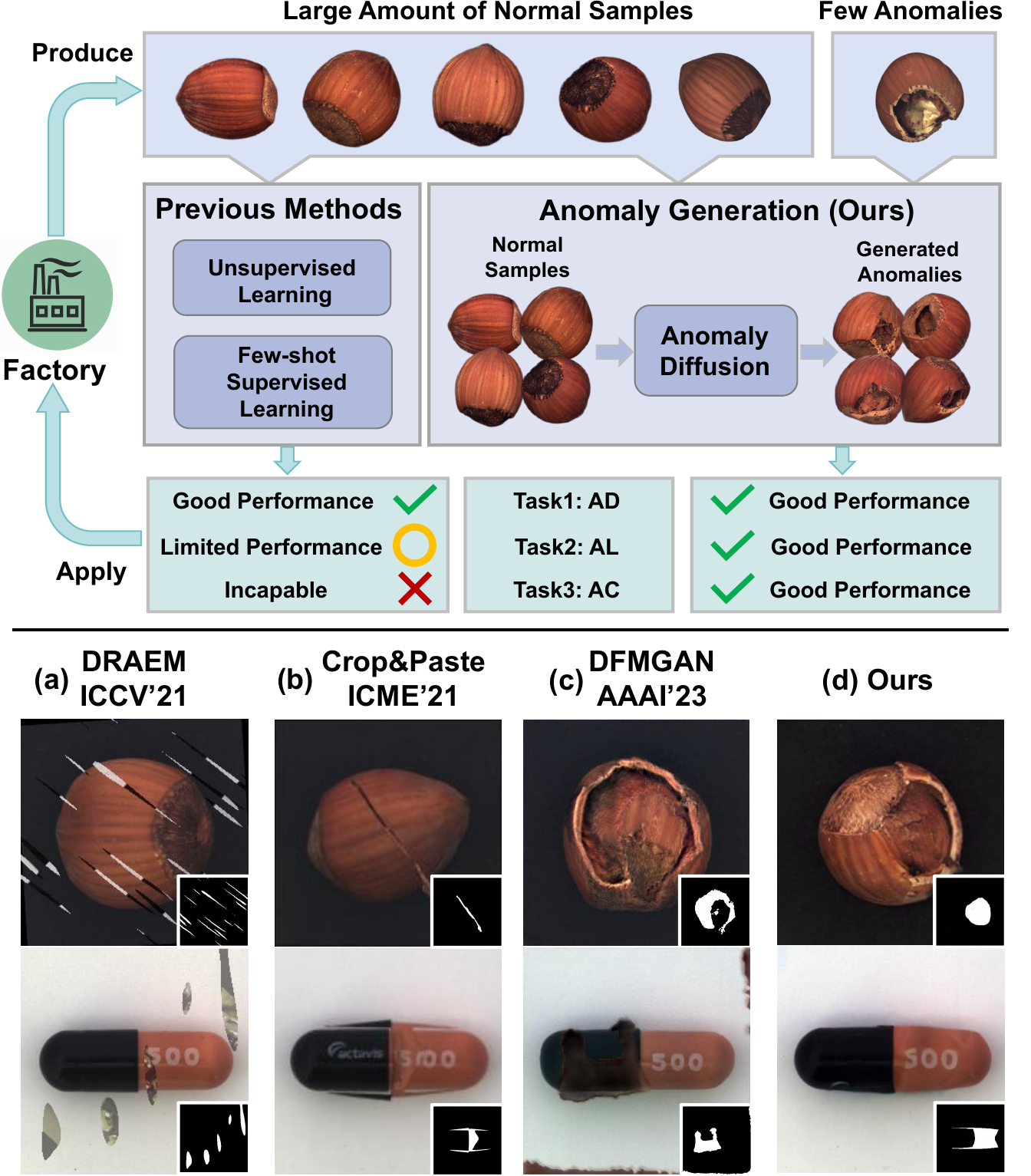}

\caption{\textbf{\textit{Top:}} 
Our model generates extensive anomaly data, which supports the downstream Anomaly Detection (AD), Localization (AL) and Classification (AC) tasks, while previous methods mainly rely on unsupervised learning or few-shot supervised learning due to the limited anomaly data; \textbf{\textit{Bottom:}} \hut{Generated anomaly results on hazelnut-crack and capsule-squeeze of} our model and existing anomaly generation methods, where our results are the most authentic.}
\label{fig:motivation}

\end{figure}

In recent years, industrial anomaly inspection algorithms, \yr{{\it i.e.,}} anomaly detection, localization, and classification, 
plays a crucial role in industrial manufacture~\cite{duan2023DFMGAN}. However, 
in real-world industrial 
\yr{production, the anomaly samples are very few,}
posing a significant challenge for anomaly inspection (Fig.~\ref{fig:motivation}-top). To mitigate the issue of 
\yr{few}
anomaly data, existing anomaly inspection mostly relies on \hut{unsupervised learning \yr{methods that} only \yr{use} normal samples~\cite{zavrtanik2021draem,li2021cutpaste}, or \yr{few-shot} supervised learning methods
~\cite{zhang2023prn}.}
\hut{Although these methods perform well in anomaly detection,} 
\yr{they have limited performance in anomaly localization and cannot handle anomaly classification.}

To cope with the 
problem of scarce anomaly samples, researchers propose \yr{anomaly generation methods to supplement the anomaly data, which can be divided into two types:} 
{ \it \textbf{1)} The model-free methods} randomly crop and paste 
\yr{patches from existing anomalies or anomaly texture dataset}
onto normal samples~\cite{li2021cutpaste,lin2021croppaste,zavrtanik2021draem}. But such methods exhibit poor authenticity in the synthesized data 
(Fig.~\ref{fig:motivation}-bottom-a/b). {\it \textbf{2)} The GAN-based methods}~\cite{zhang2021defectgan,niu2020sdgan,duan2023DFMGAN} utilize Generative Adversarial Networks (GANs)~\cite{goodfellow2014gan} to generate anomalies, but most of them 
\yr{require a large amount of}
anomaly samples for training. The only few-shot generation model DFMGAN~\cite{duan2023DFMGAN} employs StyleGAN2~\cite{karras2020stylegan2} pretrained on 
normal samples, and then performs domain adaption with a few anomaly samples. 
\hut{But the generated anomalies 
\yr{are not accurately aligned with}
the anomaly masks (Fig.~\ref{fig:motivation}-bottom-c). 
\yr{To sum up,}
the existing anomaly generation methods either fail to generate authentic anomalies or \yr{accurately}-aligned anomalous image-mask pairs by learning from few-shot anomaly data, which limits their improvement in the downstream anomaly inspection tasks.} 



\hut{
\yr{To address the above issues,}
we propose {\it AnomalyDiffusion}, a novel anomaly generation method based on the diffusion model, which generates anomalies onto the input normal samples with the anomaly masks.}
By leveraging the \yr{strong} prior information \yr{of a pretrained LDM~\cite{rombach2022ldm} learned from large-scale dataset~\cite{schuhmann2021laion}}, we can extract 
\yr{better anomaly representation}
using only a few anomaly images and boost the generation authenticity and diversity. 
\yr{To generate anomalies with specified type and locations,}
we propose \textit{Spatial Anomaly Embedding}, 
\yr{which disentangles anomaly information into}
an anomaly embedding (a learned textual embedding to represent 
\yr{the appearance} type of anomaly) and a spatial embedding 
\yr{(encoded from an anomaly mask to indicate the locations)}.
\yr{By disentangling anomaly location from appearance}, we can generate 
anomalies in any desired positions, which enables 
producing a large amount of anomalous image-mask pairs for the downstream tasks.
Moreover, we propose an \textit{Adaptive Attention Re-weighting Mechanism} to 
allocate more attention to the \hut{areas with less noticeable generated anomalies}, which dynamically adjusts the cross-attention maps based on disparities between \hut{the generated} images and input normal samples during the diffusion inference stage. 
This 
\yr{adaptive mechanism}
results in accurately aligned generated anomaly images and anomaly masks, which greatly facilitates downstream anomaly localization tasks.


Extensive qualitative and quantitative experiments and comparisons demonstrate that our {\it AnomalyDiffusion} outperforms 
\yr{state-of-the-art} anomaly generation models in terms of generation authenticity and diversity. Moreover, our generated anomaly images can be effectively applied \yr{to} downstream anomaly inspection tasks, \hut{yielding a pixel-level \textbf{99.1\%} AUROC and 
\textbf{81.4\%} AP score \yr{in anomaly localization on MVTec~\cite{bergmann2019mvtec}.}}
The main contribution of this paper can be summarized as follows:
\begin{itemize}
\item We propose {\it AnomalyDiffusion}, a few-shot diffusion-based anomaly generation method, 
which disentangles anomalies into anomaly embedding (for anomaly appearance) and spatial embedding (for anomaly location), 
and generates authentic and diverse anomaly images.

\item We design {\it Adaptive Attention Re-weighting Mechanism}, which adaptively allocates more attention to the areas \hut{with less noticeable generated anomalies}, improving the alignment between the generated anomalies and masks.


\item Extensive experiments demonstrate the superiority of our model over the state-of-the-art competitors, 
\yr{and our generated anomaly data}
effectively improves the performance of downstream anomaly inspection tasks, which will be released to 
\yr{facilitate future research.}
\end{itemize}

\section{Related Work}



\subsection{Generative Models}

{\bf Generative models.} VAEs~\cite{kingma2013vae} and GANs~\cite{goodfellow2014gan} have achieved great progress in image generation. Recently, diffusion model~\cite{nichol2021improveddif} demonstrates a more enhanced potential in generating images in a wide range of domains. Latent diffusion model (LDM)~\cite{rombach2022ldm} further improves the generation ability through compression of the diffusion space and obtains strong prior information by training on LAION dataset~\cite{schuhmann2021laion}. 

\noindent{\bf Few-shot image generation.}
Few-shot image generation aims to generate diverse images with limited training data. Early methods propose modifying network weights~\cite{mo2020freeze}, using various regularization techniques~\cite{li2020few} and data augmentation~\cite{tran2021data} to prevent overfitting. To deal with the extremely limited data (less than 10), recent works ~\cite{ojha2021few,wang2022ctlgan,hu2023few-shotdiffusion} introduce cross-domain consistency losses to keep the generated distribution. Textual Inversion ~\cite{gal2022textinversion} and Dreambooth~\cite{ruiz2023dreambooth} encode a few images into the textual space of a pre-trained LDM, but cannot control the generated locations accurately.

\subsection{Anomaly Inspection}

{\bf Anomaly inspection.}
The anomaly inspection task consists of anomaly detection, localization and classification. 
Some existing methods ~\cite{schlegl2017unsupervised,schlegl2019f,liang2023omni} rely on image reconstruction, comparing the differences between reconstructed images and anomaly images to achieve anomaly detection and localization. 
Moreover, deep feature modeling-based methods~\cite{lee2022cfa,cao2022informative,roth2022patchcore,memkd,m3dm} build a feature space for input images and then compare the differences between features to detect and localize anomalies. Additionally, some supervised learning-based methods~\cite{zhang2023prn} utilize a small number of anomaly samples to enhance the anomaly localization capabilities. Some studies conduct zero-/few-shot AD without using or with only a small number of anomaly samples~\cite{winclip,saa,aprilgan,clipad,gpt-4v-ad,regad}. Although these methods have shown promising results in anomaly detection, their performance in anomaly localization is still limited due to the lack of anomaly data.
\begin{figure*}[t]
\centering
\includegraphics[width=0.8\textwidth]{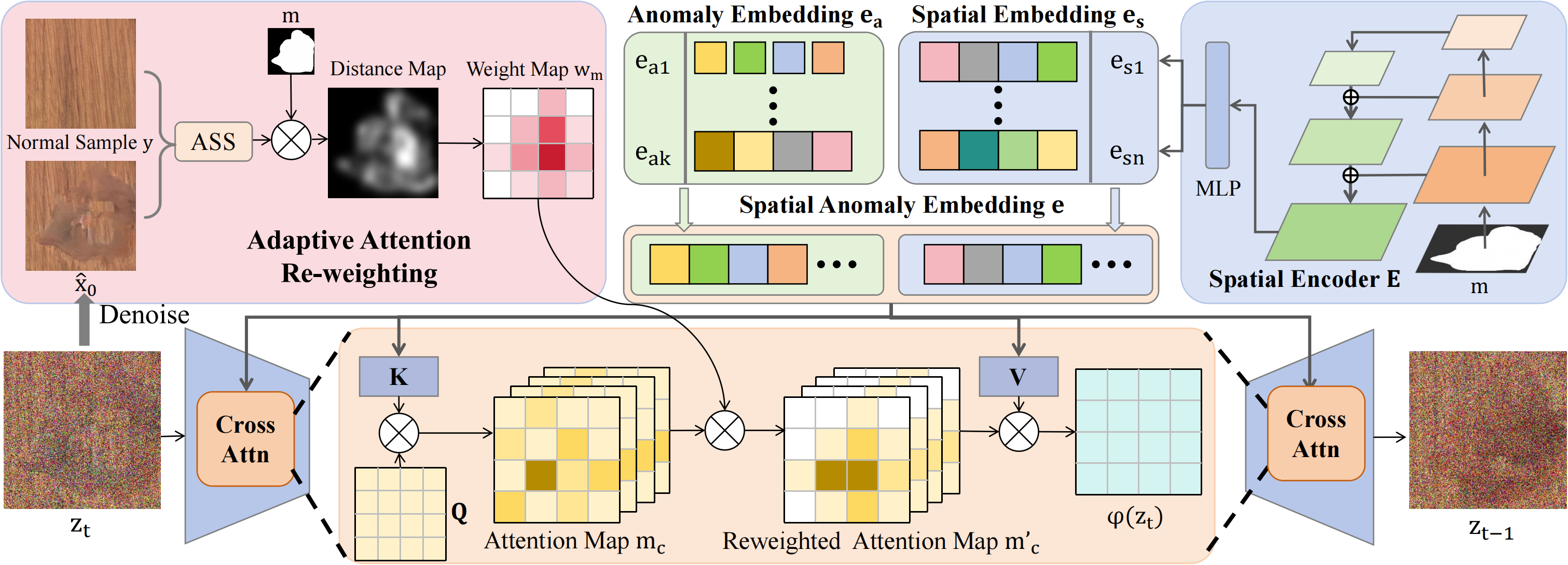}
\caption{\textbf{Overall framework of our AnomalyDiffusion:} \textbf{\textit{1)}} The {\it Spatial Anomaly Embedding} $e$, consisting of an anomaly embedding $e_a$ \yr{(a learned textual embedding to represent anomaly appearance type)} and a spatial embedding $e_s$ 
\yr{(encoded from an input anomaly mask $m$ to indicate anomaly locations),}
serves as \yr{the} text condition to guide the anomaly generation process; 
\textbf{\textit{2)}} The {\it Adaptive Attention Re-weighting Mechanism} 
computes the weight map $w_m$ \yr{based on} the difference between the denoised image $\hat{x}_0$ and the input normal sample $y$, and 
\yr{adaptively reweights}
the cross-attention map $m_c$ by the weight map $w_m$ to help the model focus more on the less noticeable anomaly areas during the denoising process. }
\label{fig:framework}
\end{figure*}

\noindent{\bf Anomaly generation.}
The scarcity of anomaly data has sparked research interest in anomaly generation. DRAEM~\cite{zavrtanik2021draem}, Cut-Paste~\cite{li2021cutpaste}, Crop-Paste~\cite{lin2021croppaste} and PRN~\cite{zhang2023prn} crop and paste unrelated textures or existing anomalies into normal sample. But they either generate less realistic anomalies or have limited generated diversity. The GAN-based model SDGAN~\cite{niu2020sdgan} and Defect-GAN~\cite{zhang2021defectgan}, generate anomalies on normal samples by learning from anomaly data. But they require a large amount of anomaly data and cannot generate anomaly mask. DFMGAN~\cite{duan2023DFMGAN} transfers a  StyleGAN2~\cite{karras2020stylegan2} pretained on normal samples to anomaly domain, but lacks generation authenticity and accurate alignment between generated anomalies and masks.
In contrast, our model incorporates spatial anomaly embedding and adaptive attention re-weighting mechanism, which can generate anomalous image-mask pairs with great diversity and authenticity.

\section{Method}







Our {\it AnomalyDiffusion} aims to generate 
\yr{a large amount of}
anomaly data \yr{aligned with anomaly masks,} 
\yr{by learning from a few}
anomaly samples. \yr{The} inputs to \yr{our model include} an \yr{anomaly-free} sample $y$ and an anomaly mask $m$, and \yr{the} output \yr{is} an image with anomalies generated in the mask area, 
\yr{while the remaining region is} consistent \yr{with} the \yr{input anomaly-free} sample. 


\hut{\yr{As shown in Fig.~\ref{fig:framework},}
our {\it AnomalyDiffusion} is developed based on Latent Diffusion Model~\cite{rombach2022ldm}. 
To disentangle the anomaly location information from anomaly appearance, we propose Spatial Anomaly Embedding $e$, which consists of an anomaly embedding $e_a$ (for anomaly appearance) and a spatial embedding $e_s$ (for anomaly location). Moreover, to enhance the alignment between the generated anomalies and given masks, we introduce an Adaptive Attention Re-weighting Mechanism, which helps the model 
\yr{to allocate more attention to}
the areas with less noticeable generated anomalies (Fig.~\ref{fig:ablation on AAR}(c)). 
}



Specifically, 
\yr{the anomaly embedding $e_a$ provides the anomaly appearance type information, with one $e_a$ corresponding to a certain type of anomaly 
({\it e.g.,} hazelnut-crack, capsule-squeeze), which is learned by our masked textual inversion (Sec.~\ref{ssec:embedding}).}
\yr{And the spatial embedding $e_s$ provides the anomaly location information, which is encoded from the input anomaly mask $m$ by a spatial encoder $E$ (shared among all anomalies).}
By combining the anomaly embedding $e_a$ with spatial embedding $e_s$, the spatial anomaly embedding $e$ contains both the anomaly appearance and spatial information, \yr{which serves as the text condition in the diffusion model to guide the generation process.} 
\yr{With the the spatial anomaly embedding as condition, given a normal sample}, we 
generate an anomaly image \hut{with the blended diffusion process~\cite{avrahami2022blended}}:
\begin{equation}
x_{t-1}= p_\theta(x_{t-1}|x_t,e) \odot m +q(y_{t-1}|y_0)\odot(1-m),
\label{eq:blended generation}
\end{equation}
where \yr{$x_t$ is the generated anomaly image at timestep $t$, $y_0$ is the input normal sample, $m$ is the anomaly mask, and}
$q(\cdot)$ and $p_{\theta}(\cdot)$ are the forward and backward process in diffusion as illustrated in Sec.~\ref{sec:preliminaries}. 

\subsection{Preliminaries}
\label{sec:preliminaries}
Denoising diffusion probabilistic models (DDPM)~\cite{ho2020ddpm} has achieved significant success in image generation tasks. It employs a forward process to add noise into the 
\yr{data}
and then learns denoising during the backward process, thereby accomplishing the fitting of the training data distribution. With the training image $x_0$, the forward process $q(\cdot)$ in diffusion model is formulated as:
\begin{equation}
   \begin{aligned}
        &q\left(x_{1}, \ldots, x_{T} \mid x_{0}\right)  :=\prod_{t=1}^{T} q\left(x_{t} \mid x_{t-1}\right), \\
        &q\left(x_{t} \mid x_{t-1}\right):=\mathcal{N}\left(x_{t} ; \sqrt{1-\beta_{t}} x_{t-1}, \beta_{t} \mathbf{I}\right),
    \end{aligned} 
\end{equation}
\yr{where $\beta_{t}$ is the variance at timestep $t$.}

The backward process is approximated by predicting the mean $\mu_{\theta}(x_t,t)$ and variance $\Sigma_{\theta}\left(x_{t}, t\right)$ (set as a constant in DDPM) of a Gaussian distribution iteratively by:
\begin{equation}
p_{\theta}\left(x_{t-1} \mid x_{t}\right) :=\mathcal{N}\left(x_{t-1} ; \mu_{\theta}\left(x_{t}, t\right), \Sigma_{\theta}\left(x_{t}, t\right)\right).
\end{equation}

Textual inversion~\cite{gal2022textinversion} utilizes a pre-trained Latent Diffusion Model 
to extract the shared content information in few-shot input samples by optimizing text embeddings. With the refined text embeddings as condition $c$, textual inversion can generate novel images $x_0$ with similar contents of input images by:
\begin{equation}
    x_0=\prod\limits_{t=1}^T p_{\theta}(x_{t-1}|x_{t},c), \, x_T\sim \mathcal{N}(0,1).
\end{equation}

\subsection{Spatial Anomaly Embedding}
\label{ssec:embedding}

\hut{\noindent\textbf{Disentangle spatial information from anomaly appearance.}  
\yr{We aim at controllable anomaly generation with specified anomaly type and location.
A direct solution is to control anomaly type by textual embedding learned from textual inversion~\cite{gal2022textinversion}, and control anomaly location by the input mask.}
However, textual inversion tends to 
\yr{capture the location of anomalies along with the anomaly type information,}
\yr{which results in the generated anomalies only distributed in specific locations.}}
\yr{To address the issue, we propose to disentangle the textual embedding into two parts, where one part (the spatial embedding $e_s$) is directly encoded from the anomaly mask to indicate the anomaly location, leaving the rest (the anomaly embedding $e_a$) to only learn anomaly type information. We name our decomposed textual embedding as Spatial Anomaly Embedding.}



\noindent\textbf{Anomaly embedding}
\yr{is a learned textual embedding that represents the anomaly appearance type information.}
\yr{Different from textual inversion method that learns the features of the entire image,}
in anomaly generation, our model only needs to focus on anomaly areas, without requiring information of the entire image. 
Therefore, we introduce {\it masked textual inversion}, where we mask out irrelevant background and normal regions of the anomaly image, and only the anomaly regions are visible to the model.
We initialize the anomaly embedding $e_a$ with $k$ tokens and optimize it using the masked diffusion loss:
\begin{equation}
\begin{aligned}
\mathcal{L}_{dif}=\left\| m \odot(\epsilon-\epsilon_\theta\left(z_t, t, \{ e_a, e_s \}\right))\right\|_2^2,
\end{aligned}
\label{eq:mased textual inversion}
\end{equation}
where $\epsilon \sim \mathcal{N}(0,1)$ and $z_t$ is the \yr{noised} latent code of the input image $x$ 
\yr{at timestep $t$.}

\noindent\textbf{Spatial embedding.}
To provide accurate spatial information 
\yr{of the anomaly locations,}
we introduce a spatial encoder $E$ that encodes the input anomaly mask $m$ into spatial embedding $e_s$, which is 
in the form of textual embedding and contains precise location information from the mask. 
Specifically, we \yr{input the anomaly mask into} ResNet-50~\cite{he2016resnet} to extract the image features in different layers and fuse them together by Feature Pyramid Networks~\cite{lin2017fpn}. Finally, several fully-connected networks are employed to map the fused features into textual \yr{embedding} space, with each network \yr{predicting} one text token, thereby outputting the final spatial embedding $e_s$ with $n$ tokens.

\noindent\textbf{Overall training framework.} For each anomaly \yr{type} $i$, we employ an anomaly embedding $e_{a,i}$ to extract its appearance information, while all anomaly categories share a common spatial encoder $E$. For a set of image-mask pairs $(x_i, m_i)$ in the training data, we first input anomaly mask $m_i$ into spatial encoder $E$ to obtain the spatial embedding $e_s=E(m_i)$. Then, we concatenate the anomaly embedding $e_{a,i}$ and the spatial embedding $e_s$ together to obtain our spatial anomaly embedding $e=\{e_a,e_s\}$. Finally, the concatenated textual embedding $e$ is used as the text condition to the diffusion model, 
and the training process can be formulated as:
\begin{equation}
\begin{aligned}
e_a^*,E^*=&\mathop{\arg\min}\limits_{e_a,E}  \mathbb{E}_{z \sim \mathcal{E}(x_i), m_i, \epsilon, t} \mathcal{L}_{dif}.
\end{aligned}
\label{eq:overall training framework}
\end{equation}
where $\mathcal{E}(\cdot)$ is the image encoder of latent diffusion model and $\epsilon \sim \mathcal{N}(0,1)$.

\begin{figure}[t]
\centering
\includegraphics[width=0.36\textwidth]{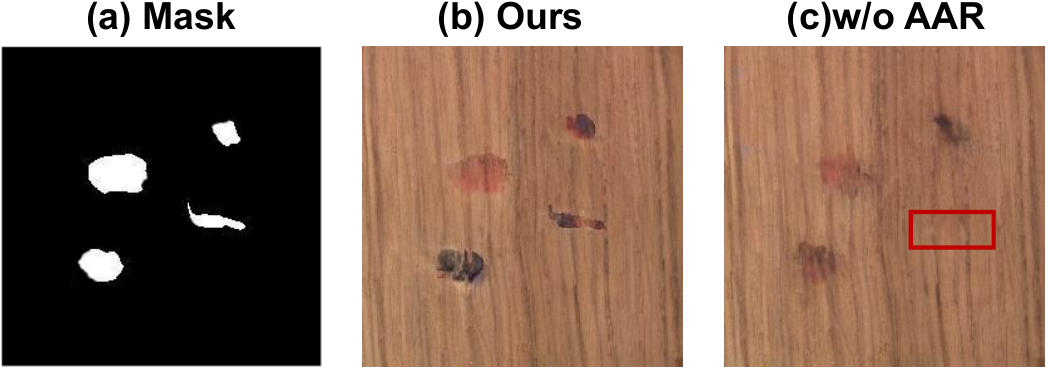}

\caption{Comparison between the models w/ (Ours) and w/o Adaptive Attention Re-weighting (AAR). The model w/o AAR cannot generate anomalies to fill 
the entire mask.}
\label{fig:ablation on AAR}

\end{figure}

\subsection{Adaptive Attention Re-Weighting}



With the spatial anomaly embedding $e$, we can use it as the text condition to guide the generation of anomaly images by Eq.~(\ref{eq:blended generation}). However, the generated anomaly images sometimes fail to fill the entire mask, especially when there are multiple anomaly regions in the mask or when the mask has irregular shapes (Fig.~\ref{fig:ablation on AAR}-a/c). In such cases, the generated anomalies are usually not well aligned with the mask, 
\yr{which limits the improvement in downstream anomaly localization task.}
To address this problem, we propose an adaptive attention re–weighting mechanism, which 
\yr{allocates more attention to the}
areas with less noticeable generated anomalies during the denoising process, thereby facilitating better alignment between the generated anomalies and the anomaly masks.

\noindent\textbf{Adaptive attention weight map.} Specifically, at the $t$-th denoising step, we calculate the corresponding $\hat{x}_0=D(p_\theta(\hat{z}_0|z_t,e))$ (where $D$ is the decoder of LDM). Then, we \yr{calculate the pixel-level difference between} 
$\hat{x}_0$ and the normal sample $y$ within the mask $m$. 
Based on the difference, we calculate the 
weight map $w_m$ \hut{by the Adaptive Scaling Softmax (ASS) operation}:
\begin{equation}
\begin{aligned}
w_m=\|m\|_1\cdot Softmax(f(\|m\odot y-m\odot \hat{x}_0\|^2_2)),
\end{aligned}
\label{eq:attention weight map}
\end{equation}
where $f(x)=\frac{1}{x}$ when $x!=0$ and $f(x)=-\infty$ otherwise.
\hut{For the regions within the mask that are similar to normal samples, the generated anomalies \yr{in these regions} are less noticeable. 
\yr{To enhance the anomaly generation effects}, 
these regions are assigned higher weights by Eq.~(\ref{eq:attention weight map})
\yr{and allocated with more attention by attention re-weighting.}
}
\begin{table*}[t!]
\centering
\renewcommand{\arraystretch}{1.0}
\setlength\tabcolsep{12pt}
\resizebox{1.0\linewidth}{!}{
\begin{tabular}{p{2.5cm}<{\centering}|cc|cc|cc|cc|cc|cc|cc}
\toprule
\multirow{2}{*}{Category} &
  \multicolumn{2}{c}{DiffAug} &
  \multicolumn{2}{c}{CDC} &
  \multicolumn{2}{c}{Crop-Paste} &
  \multicolumn{2}{c}{SDGAN} &
  \multicolumn{2}{c}{Defect-GAN} &
  \multicolumn{2}{c}{DFMGAN} &
  \multicolumn{2}{c}{Ours} \\
 &
  IS $\uparrow$ &
 IC-L $\uparrow$ &
  IS $\uparrow$ &
 IC-L $\uparrow$ &
  IS $\uparrow$ &
 IC-L $\uparrow$ &
  IS $\uparrow$ &
 IC-L &
  IS $\uparrow$ &
 IC-L $\uparrow$ &
  IS $\uparrow$ &
 IC-L $\uparrow$ &
  IS $\uparrow$ &
 IC-L $\uparrow$ \\
  \toprule
bottle & \underline{1.59} & 0.03 & 1.52 & 0.04 & 1.43 & 0.04 & 1.57 & 0.06 & 1.39 & 0.07 & \textbf{1.62} & \underline{0.12} & 1.58 & \textbf{0.19} \\
cable & 1.72 & 0.07 & \underline{1.97} & 0.19 & 1.74 & \underline{0.25} & 1.89 & 0.19 & 1.70 & 0.22 & 1.96 & \underline{0.25} & \textbf{2.13} & \textbf{0.41} \\
capsule & 1.34 & 0.03 & 1.37 & 0.06 & 1.23 & 0.05 & \underline{1.49} & 0.03 & \textbf{1.59} & 0.04 & \textbf{1.59} & \underline{0.11} & \textbf{1.59} & \textbf{0.21} \\
carpet & 1.19 & 0.06 & \textbf{1.25} & 0.03 & 1.17 & 0.11 & 1.18 & 0.11 & \underline{1.24} & 0.12 & 1.23 & \underline{0.13} & 1.16 & \textbf{0.24} \\
grid & 1.96 & 0.06 & 1.97 & 0.07 & 2.00 & 0.12 & 1.95 & 0.10 & \underline{2.01} & 0.12 & 1.97 & \underline{0.13} & \textbf{2.04} & \textbf{0.44} \\
hazel\_nut & 1.67 & 0.05 & \underline{1.97} & 0.05 & 1.74 & 0.21 & 1.85 & 0.16 & 1.87 & 0.19 & 1.93 & \underline{0.24} & \textbf{2.13} & \textbf{0.31} \\
leather & \underline{2.07} & 0.06 & 1.80 & 0.07 & 1.47 & 0.14 & 2.04 & 0.12 & \textbf{2.12} & 0.14 & 2.06 & \underline{0.17} & 1.94 & \textbf{0.41} \\
metal nut & \underline{1.58} & 0.29 & 1.55 & 0.04 & 1.56 & 0.15 & 1.45 & 0.28 & 1.47 & \underline{0.30} & 1.49 & \textbf{0.32} & \textbf{1.96} & \underline{0.30} \\
pill & 1.53 & 0.05 & 1.56 & 0.06 & 1.49 & 0.11 & \underline{1.61} & 0.07 & \underline{1.61} & 0.10 & \textbf{1.63} & \underline{0.16} & \underline{1.61} & \textbf{0.26} \\
screw & 1.10 & 0.10 & 1.13 & 0.11 & 1.12 & \underline{0.16} & 1.17 & 0.10 & \underline{1.19} & 0.12 & 1.12 & 0.14 & \textbf{1.28} & \textbf{0.30} \\
tile & 1.93 & 0.09 & 2.10 & 0.12 & 1.83 & 0.20 & \underline{2.53} & 0.21 & 2.35 & \underline{0.22} & 2.39 & \underline{0.22} & \textbf{2.54} & \textbf{0.55} \\
toothbrush & 1.33 & 0.06 & 1.63 & 0.06 & 1.30 & 0.08 & 1.78 & 0.03 & \textbf{1.85} & 0.03 & \underline{1.82} & \underline{0.18} & 1.68 & \textbf{0.21} \\
transistor & 1.34 & 0.05 & 1.61 & 0.13 & 1.39 & 0.15 & \textbf{1.76} & 0.13 & 1.47 & 0.13 & \underline{1.64} & \underline{0.25} & 1.57 & \textbf{0.34} \\
wood & 2.05 & 0.30 & 2.05 & 0.03 & 1.95 & 0.23 & 2.12 & 0.25 & \underline{2.19} & 0.29 & 2.12 & \underline{0.35} & \textbf{2.33} & \textbf{0.37} \\
zipper & \underline{1.30} & 0.05 & \underline{1.30} & 0.05 & 1.23 & 0.11 & 1.25 & 0.10 & 1.25 & 0.10 & 1.29 & \textbf{0.27} & \textbf{1.39} & \underline{0.25} \\
\midrule
Average & 1.58 & 0.09 & 1.65 & 0.07 & 1.51 & 0.14 & 1.71 & 0.13 & 1.69 & 0.15 & \underline{1.72} & \underline{0.20} & \textbf{1.80} & \textbf{0.32}\\
\bottomrule
\end{tabular}}
\caption{\textbf{Comparison on IS and IC-LPIPS on MVTec dataset.} Our model generates the most high-quality and diverse anomaly data, achieving the best IS and IC-LPIPS. \textbf{Bold} and \underline{underline} represent optimal and sub-optimal results, respectively.}
\label{tab:comparison on kid and IC-LPIPS}
\end{table*}

\noindent\textbf{Attention re-weighting.} We employ the weight map $w_m$ to \yr{adaptively} control the cross-attention, \yr{in order to guide} our model to focus more on the areas with less noticeable generated anomalies. 
\yr{In our cross-attention calculation, Query is calculated from the latent code $z_t$, and Key and Value are calculated from our spatial anomaly embedding $e$: }
\begin{equation}
Q=W_Q^{(i)} \cdot \varphi_i\left(z_t\right), K=W_K^{(i)} \cdot e, V=W_V^{(i)} \cdot e,
\label{eq:cross attention}
\end{equation}
where $\varphi_i$ is the intermediate representation of the U-Net ($\epsilon_\theta$) and the $W^{(i)}$s are the learnable projection matrices.
The cross-attention calculation process is then formulated as $Attn(Q,K,V)=m_c\cdot V$, \yr{where $m_c=Softmax(\frac{QK^T}{\sqrt{d}})$ is the cross-attention map.}

\begin{figure}[t!]
\centering
\includegraphics[width=0.4\textwidth]{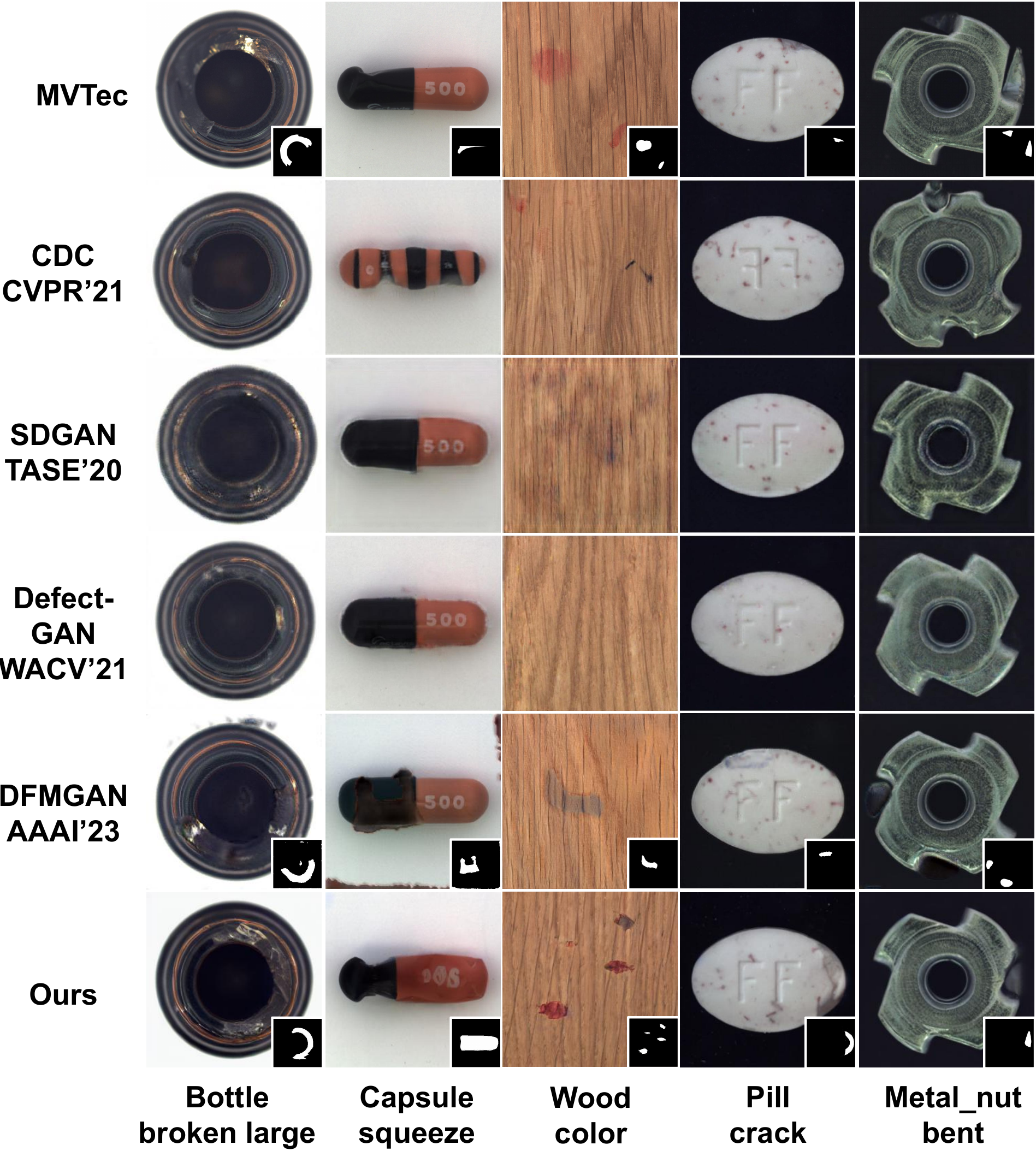}

\caption{\textbf{Comparison on the generation results on MVTec.} Our model \yr{generates high quality} anomaly \yr{images that are accurately aligned with the anomaly masks}. 
}
\label{fig:generation quality}

\end{figure}

Considering the cross-attention map $m_c$ controls the generated layout 
\yr{and effects, where higher attention leads to stronger generation effects}~\cite{hertz2022prompt}, we reweight the cross-attention map by our weight map: $m'_c=m_c \odot w_m$. 
The new cross-attention map $m'_c$ focuses more on the areas with less noticeable \yr{generated} anomalies, thereby enhancing the alignment accuracy between the generated anomalies and the input anomaly masks. The re-weighted cross attention 
\yr{is formulated as $RW\text{-}Attn(Q,K,V)=m'_c \cdot V.$} 


\subsection{Mask Generation}

\yr{Recall that our model requires anomaly masks as inputs. 
However, the number of real anomaly masks in the training datasets is very few, and the mask data lacks diversity even after augmentation, which motivates us to generate more anomaly masks by learning the real mask distribution.}
\hut{
We employ textual inversion to learn a mask embedding $e_m$, 
which can be used as text condition to generate extensive anomaly masks.} 
Specifically, we initialize the mask embedding $e_m$ as $k'$ random tokens and optimize it by:
\begin{equation}
e_m^*=\mathop{\arg\min}\limits_{e_m}  \mathbb{E}_{z \sim \mathcal{E}(m), \epsilon , t}\left[\left\|\epsilon-\epsilon_\theta\left(z_t, t, e_m\right)\right\|_2^2\right].
\end{equation}
With the learned mask embedding, we can generate extensive anomaly masks for each type of anomaly.

\section{Experiments}

\subsection{Experiment Settings}

\textbf{Dataset.} 
we conduct experiments on the widely used MVTec~\cite{bergmann2019mvtec} dataset.
 We employ one-third of the anomaly data with the lowest ID numbers as the training set, reserving the remaining two-thirds for testing.

\noindent\textbf{Implementation details.} We assign $k=8$ tokens for anomaly embedding, $n=4$ tokens for spatial embedding, and $k'=4$ tokens for mask embedding. For each type of anomaly, we generate 1000 anomalous image-mask pairs for the downstream anomaly inspection tasks. More details are recorded in the supplementary material.

\noindent\textbf{Metric.}
{\bf \textit{1)} For generation,} 
due to the limited anomaly data, FID~\cite{heusel2017fid} and KID~\cite{binkowski2018kid} are not reliable since the overfitted model tends to yield better scores (best)~\cite{duan2023DFMGAN}. Therefore, we employ Inception Score \textbf{(IS)}, which is independent of the given anomaly data, for a direct assessment of generation quality; 
 we also introduce Intra-cluster pairwise LPIPS distance (\textbf{IC-LPIPS})~\cite{ojha2021few} to measure the generation diversity.
{\bf \textit{2)} for anomaly inspection,} 
 we utilize \textbf{AUROC}, Average Precision (\textbf{AP}), and the \textbf{$\mathbf{F_1}$-max} score to evaluate the accuracy of anomaly detection and localization. 

\begin{figure}[t!]
\centering
\includegraphics[width=0.38\textwidth]{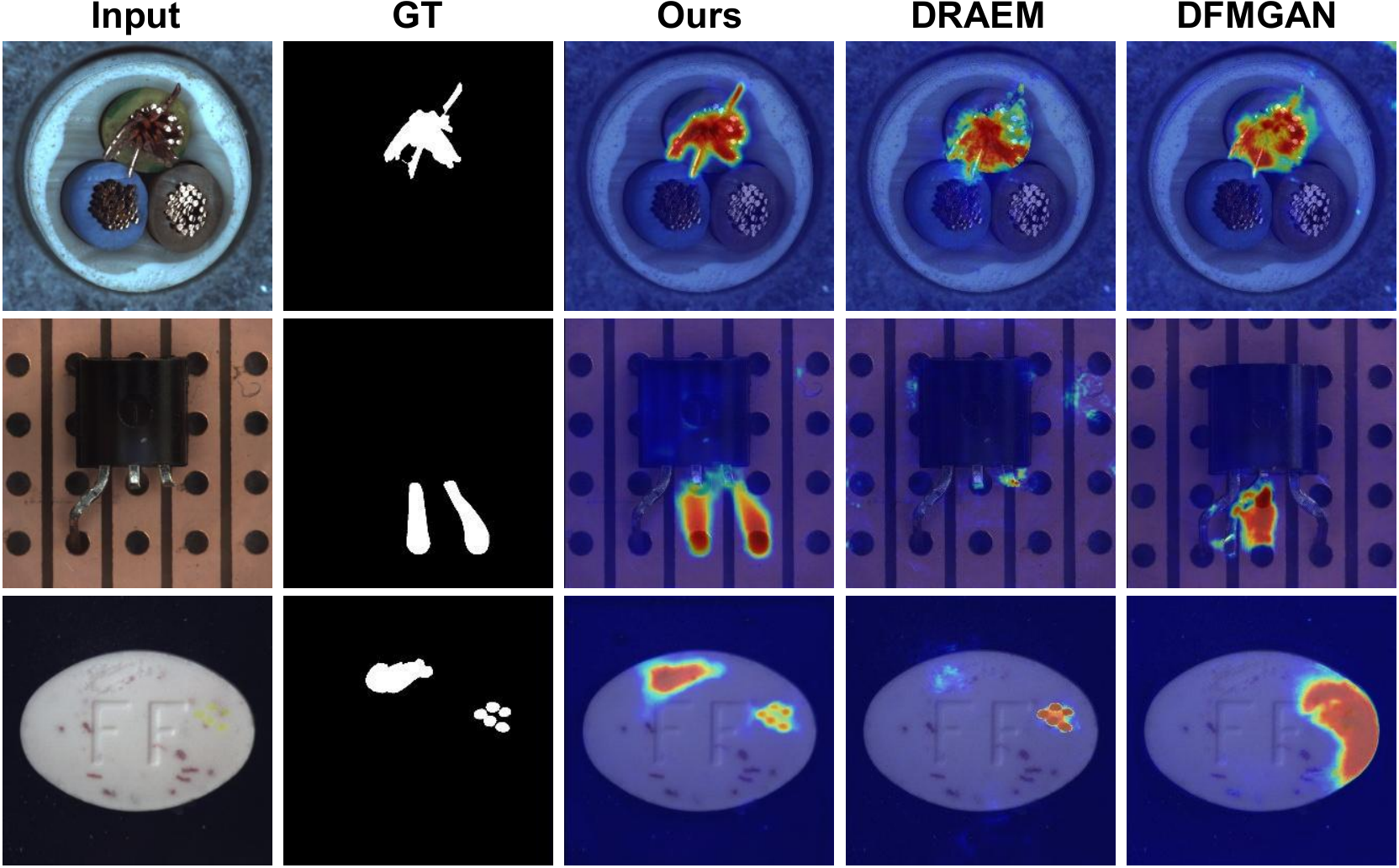}

\caption{\textbf{Quantitative anomaly localization comparison} with an U-Net trained on the data generated by DRAEM, DFMGAN and our model. It shows that our model achieves the best anomaly localization results. }
\label{fig: qualitative comparison}

\end{figure}

\begin{table}[t!]
\centering
\renewcommand{\arraystretch}{1.0}
\setlength\tabcolsep{5.0pt}
\resizebox{1.0\linewidth}{!}{
\begin{tabular}{c|ccc|ccc|ccc|ccc}
\toprule
\multirow{2}{*}{Category} & \multicolumn{3}{c}{DRAEM} & \multicolumn{3}{c}{PRN} & \multicolumn{3}{c}{DFMGAN} & \multicolumn{3}{c}{Ours} \\
 & AUC & AP & $F_1$-max & AUC & AP & $F_1$-max & AUC & AP & $F_1$-max & AUC & AP & $F_1$-max \\
 \midrule
bottle & 96.7 & 80.2 & 74.0 & 97.5 & 76.4 & 71.3 & \underline{98.9} & \underline{90.2} & \underline{83.9} & \textbf{99.4} & \textbf{94.1} & \textbf{87.3} \\
cable & 80.3 & 21.8 & 28.3 & 94.5 & 64.4 & 61.0 & \underline{97.2} & \underline{81.0} & \underline{75.4} & \textbf{99.2} & \textbf{90.8} & \textbf{83.5} \\
capsule & 76.2 & 25.5 & 32.1 & \underline{95.6} & \underline{45.7} & \underline{47.9} & 79.2 & 26.0 & 35.0 & \textbf{98.8} & \textbf{57.2} & \textbf{59.8} \\
carpet & 92.6 & 43.0 & 41.9 & \underline{96.4} & \underline{69.6} & \underline{65.6} & 90.6 & 33.4 & 38.1 & \textbf{98.6} & \textbf{81.2} & \textbf{74.6} \\
grid & \textbf{99.1} & \textbf{59.3} & \underline{58.7} & \underline{98.9} & \underline{58.6} & \textbf{58.9} & 75.2 & 14.3 & 20.5 & 98.3 & 52.9 & 54.6 \\
hazelnut & 98.8 & 73.6 & 68.5 & 98.0 & 73.9 & 68.2 & \underline{99.7} & \underline{95.2} & \underline{89.5} & \textbf{99.8} & \textbf{96.5} & \textbf{90.6} \\
leather & 98.5 & 67.6 & 65.0 & \underline{99.4} & 58.1 & 54.0 & 98.5 & \underline{68.7} & \underline{66.7} & \textbf{99.8} & \textbf{79.6} & \textbf{71.0} \\
metal nut & 96.9 & 84.2 & 74.5 & 97.9 & 93.0 & 87.1 & \underline{99.3} & \underline{98.1} & \textbf{94.5} & \textbf{99.8} & \textbf{98.7} & \underline{94.0} \\
pill & 95.8 & 45.3 & 53.0 & \underline{98.3} & 55.5 & \underline{72.6} & 81.2 & \underline{67.8} & \underline{72.6} & \textbf{99.8} & \textbf{97.0} & \textbf{90.8} \\
screw & 91.0 & 30.1 & 35.7 & \underline{94.0} & \underline{47.7} & \underline{49.8} & 58.8 & 2.2 & 5.3 & \textbf{97.0} & \textbf{51.8} & \textbf{50.9} \\
tile & 98.5 & 93.2 & \underline{87.8} & 98.5 & 91.8 & 84.4 & \textbf{99.5} & \textbf{97.1} & \textbf{91.6} & \underline{99.2} & \underline{93.9} & 86.2 \\
toothbrush & 93.8 & 29.5 & 28.4 & 96.1 & 46.4 & 46.2 & \underline{96.4} & \underline{75.9} & \underline{72.6} & \textbf{99.2} & \textbf{76.5} & \textbf{73.4} \\
transistor & 76.5 & 31.7 & 24.2 & 94.9 & 68.6 & 68.4 & \underline{96.2} & \underline{81.2} & \underline{77.0} & \textbf{99.3} & \textbf{92.6} & \textbf{85.7} \\
wood & \underline{98.8} & \textbf{87.8} & \textbf{80.9} & 96.2 & 74.2 & 67.4 & 95.3 & 70.7 & 65.8 & \textbf{98.9} & \underline{84.6} & \underline{74.5} \\
zipper & 93.4 & 65.4 & 64.7 & \underline{98.4} & \underline{79.0} & \underline{73.7} & 92.9 & 65.6 & 64.9 & \textbf{99.4} & \textbf{86.0} & \textbf{79.2} \\
\midrule
Average & 92.2 & 54.1 & 53.1 & \underline{96.9} & \underline{66.2} & \underline{64.7} & 90.0 & 62.7 & 62.1 & \textbf{99.1} & \textbf{81.4} & \textbf{76.3} \\
\bottomrule
\end{tabular}}
\caption{\textbf{Comparison on pixel-level anomaly localization on MVTec dataset} by training an U-Net on the generated data from DRAEM, PRN, DFMGAN and our model.}
\label{tab:pixel level anomaly localization comparison}
\end{table}

\subsection{Comparison in Anomaly Generation}


\textbf{Baseline.}
The compared anomaly generation methods can be classified into 2 groups: \textbf{\textit{1)}} the models (Crop\&Paste~\cite{lin2021croppaste}, DRAEM~\cite{zavrtanik2021draem}, PRN~\cite{zhang2023prn} and DFMGAN~\cite{duan2023DFMGAN}) that can generate anomalous image-mask pairs, which are employed to compare anomaly detection and localization; \textbf{\textit{2)}} the models (DiffAug~\cite{zhao2020diffaug}, CDC~\cite{ojha2021few}, Crop\&Paste, SDGAN~\cite{niu2020sdgan}, Defect-GAN~\cite{zhang2021defectgan} and DFMGAN) that can generate specific anomaly types, which are employed to compare anomaly generation quality and classification.


\begin{table}[t!]
\renewcommand{\arraystretch}{1.0}
\setlength\tabcolsep{5.0pt}
\resizebox{1.\linewidth}{!}{
\begin{tabular}{c|ccc|ccc|ccc|ccc}
\toprule
\multirow{2}{*}{Category} & \multicolumn{3}{c}{DRAEM} & \multicolumn{3}{c}{PRN} & \multicolumn{3}{c}{DFMGAN} & \multicolumn{3}{c}{Ours} \\
 & AUC & AP & $F_1$-max & AUC & AP & $F_1$-max & AUC & AP & $F_1$-max & AUC & AP & $F_1$-max \\
 \midrule
bottle & \underline{99.3} & \underline{99.8} & \textbf{98.9} & 94.9 & 98.4 & 94.1 & \underline{99.3} & \underline{99.8} & \underline{97.7} & \textbf{99.8} & \textbf{99.9} & \textbf{98.9} \\
cable & 72.1 & 83.2 & 79.2 & 86.3 & 92.0 & 84.0 & \underline{95.9} & \underline{97.8} & \underline{93.8} & \textbf{\textbf{\textbf{100}}} & \textbf{\textbf{\textbf{100}}} & \textbf{\textbf{\textbf{100}}} \\
capsule & \underline{93.2} & \underline{98.7} & 94.0 & 84.9 & 95.8 & 94.3 & 92.8 & 98.5 & \underline{94.5} & \textbf{99.7} & \textbf{99.9} & \textbf{98.7} \\
carpet & \underline{95.3} & \underline{98.7} & \underline{93.4} & 92.6 & 97.8 & 92.1 & 67.9 & 87.9 & 87.3 & \textbf{96.7} & \textbf{98.8} & \textbf{94.3} \\
grid & \textbf{99.8} & \textbf{99.9} & \textbf{98.8} & 96.6 & 98.9 & 95.0 & 73.0 & 90.4 & 85.4 & \underline{98.4} & \underline{99.5} & \underline{98.7} \\
hazelnut & \textbf{\textbf{\textbf{100}}} & \textbf{\textbf{\textbf{100}}} & \textbf{\textbf{\textbf{100}}} & 93.6 & 96.0 & 94.1 & \underline{99.9} & \textbf{\textbf{\textbf{100}}} & \underline{99.0} & 99.8 & \underline{99.9} & 98.9 \\
leather & \textbf{\textbf{\textbf{100}}} & \textbf{\textbf{\textbf{100}}} & \textbf{\textbf{\textbf{100}}} & 99.1 & \underline{99.7} & 97.6 & \underline{99.9} & \textbf{\textbf{\textbf{100}}} & \underline{99.2} & \textbf{\textbf{\textbf{100}}} & \textbf{\textbf{\textbf{100}}} & \textbf{\textbf{\textbf{100}}} \\
metal\_nut & 97.8 & 99.6 & 97.6 & 97.8 & 99.5 & 96.9 & \underline{99.3} & \underline{99.8} & \underline{99.2} & \textbf{\textbf{\textbf{100}}} & \textbf{\textbf{\textbf{100}}} & \textbf{\textbf{\textbf{100}}} \\
pill & \underline{94.4} & \underline{98.9} & \underline{95.8} & 88.8 & 97.8 & 93.2 & 68.7 & 91.7 & 91.4 & \textbf{98.0} & \textbf{99.6} & \textbf{97.0} \\
screw & \underline{88.5} & \underline{96.3} & \underline{89.3} & 84.1 & 94.7 & 87.2 & 22.3 & 64.7 & 85.3 & \textbf{96.8} & \textbf{97.9} & \textbf{95.5} \\
tile & \textbf{\textbf{\textbf{100}}} & \textbf{\textbf{\textbf{100}}} & \textbf{\textbf{\textbf{100}}} & \underline{91.1} & \underline{96.9} & \underline{89.3} & \textbf{\textbf{\textbf{100}}} & \textbf{\textbf{\textbf{100}}} & \textbf{\textbf{\textbf{100}}} & \textbf{\textbf{\textbf{100}}} & \textbf{\textbf{\textbf{100}}} & \textbf{\textbf{\textbf{100}}} \\
toothbrush & \underline{99.4} & \underline{99.8} & \underline{97.6} & \textbf{\textbf{\textbf{100}}} & \textbf{\textbf{\textbf{100}}} & \textbf{\textbf{\textbf{100}}} & \textbf{\textbf{\textbf{100}}} & \textbf{\textbf{\textbf{100}}} & \textbf{\textbf{\textbf{100}}} & \textbf{\textbf{\textbf{100}}} & \textbf{\textbf{\textbf{100}}} & \textbf{\textbf{\textbf{100}}} \\
transistor & 79.6 & 80.5 & 71.4 & 88.2 & \underline{88.9} & 84.0 & \underline{90.8} & \underline{92.5} & \underline{88.9} & \textbf{\textbf{\textbf{100}}} & \textbf{\textbf{\textbf{100}}} & \textbf{\textbf{\textbf{100}}} \\
wood & \textbf{\textbf{\textbf{100}}} & \textbf{\textbf{\textbf{100}}} & \textbf{\textbf{\textbf{100}}} & 77.5 & 92.7 & 86.7 & \underline{98.4} & \underline{99.4} & \underline{98.8} & \underline{98.4} & \underline{99.4} & \underline{98.8} \\
zipper & \textbf{\textbf{\textbf{100}}} & \textbf{\textbf{\textbf{100}}} & \textbf{\textbf{\textbf{100}}} & 98.7 & 99.7 & 97.6 & 99.7 & \underline{99.9} & \underline{99.4} & \underline{99.9} & \textbf{\textbf{\textbf{100}}} & \underline{99.4} \\
\midrule
Average & \underline{94.6} & \underline{97.0} & 94.4 & 91.6 & 96.6 & 92.4 & 87.2 & 94.8 & \underline{94.7} & \textbf{99.2} & \textbf{99.7} & \textbf{98.7}\\
\bottomrule
\end{tabular}}
\caption{\textbf{Comparison on image-level anomaly detection.}}
\label{tab:image-level anomlay detection comparison}
\end{table}

\begin{table}[t!]
\renewcommand{\arraystretch}{1.0}
\setlength\tabcolsep{5.0pt}
\resizebox{1.\linewidth}{!}{
\begin{tabular}{c|ccccccc}
\toprule
 Category& DiffAug & CDC & Crop\&Paste & SDGAN & Defect-GAN & DFMGAN & Ours \\
\midrule
bottle & 48.84 & 38.76 & 52.71 & 48.84 & 53.49 & \underline{56.59} & \textbf{90.70} \\
cable & 21.36 & 39.06 & 32.81 & 21.88 & 21.36 & \underline{45.31} & \textbf{67.19} \\
capsule & 34.67 & 28.89 & 32.89 & 30.22 & 32.00 & \underline{37.23} & \textbf{66.67} \\
carpet & 35.48 & 25.27 & 27.96 & 21.50 & 29.03 & \underline{47.31} & \textbf{58.06} \\
grid & 28.33 & 35.83 & 28.33 & 30.83 & 27.50 & \underline{40.83} & \textbf{42.50} \\
hazelnut & 65.28 & 54.86 & 59.03 & 43.75 & 61.11 & \underline{81.94} & \textbf{85.42} \\
leather & 40.74 & 43.38 & 34.39 & 38.10 & 42.33 & \underline{49.73} & \textbf{61.90} \\
metalnut & 58.85 & 48.44 & \underline{59.89} & 44.27 & 56.77 & \textbf{64.58} & 59.38 \\
pill & \underline{29.86} & 21.88 & 26.74 & 20.49 & 28.47 & 29.52 & \textbf{59.38} \\
screw & 25.10 & 32.92 & 28.81 & 26.75 & 28.81 & \underline{37.45} & \textbf{48.15} \\
tile & 59.65 & 48.54 & 68.42 & 42.69 & 26.90 & \underline{74.85} & \textbf{84.21} \\
transistor & 38.09 & 29.76 & 41.67 & 32.14 & 35.72 & \underline{52.38} & \textbf{60.71} \\
wood & 41.27 & 28.57 & 47.62 & 30.95 & 24.60 & \underline{49.21} & \textbf{71.43} \\
zipper & 22.76 & 14.63 & 26.42 & 21.54 & 18.70 & \underline{27.64} & \textbf{69.51} \\
\midrule
Average & 39.31 & 35.06 & 40.55 & 32.43 & 34.77 & \underline{49.61} & \textbf{66.09}\\
\bottomrule
\end{tabular}}
\caption{\textbf{Comparison on anomaly classification accuracy} trained on the generated data by the anomaly generation models with a ResNet-18.}
\label{tab:comparison on anomaly classfication. }
\end{table}

\begin{table*}[t!]
\renewcommand{\arraystretch}{1.0}
\setlength\tabcolsep{10.0pt}
\resizebox{1.\linewidth}{!}{
\begin{tabular}{@{}cccccccccccc@{}}
\toprule
\multirow{2}{*}{Category}     & \multicolumn{7}{c}{Unsupervised}                                               & \multicolumn{4}{c}{Supervised} \\
\cmidrule(lr){2-8}\cmidrule(lr){9-12} 
            & KDAD& CFLOW & DRAEM & SSPCAB& CFA & RD4AD& PatchCore & DevNet & DRA& PRN &Ours   \\ \midrule
bottle & 94.7/50.5 & 98.8/49.9 & 99.1/88.5 & 98.9/88.6 & 98.9/50.9 & 98.8/51.0 & 97.6/75.0 & 96.7/67.9 & 91.7/41.5 & \textbf{99.4}/92.3 & 99.3/\textbf{94.1} \\
cable & 79.2/11.6 & 98.9/72.6 & 94.8/61.4 & 93.1/52.1 & 98.4/79.8 & 98.8/77.0 & 96.8/65.9 & 97.9/67.6 & 86.1/34.8 & 98.8/78.9 & \textbf{99.2}/\textbf{90.8} \\
capsule & 96.3/\pzo 9.9 & \textbf{99.5}/64.0 & 97.6/47.9 & 90.4/48.7 & 98.9/\textbf{71.1} & 99.0/60.5 & 98.6/46.6 & 91.1/46.6 & 88.5/11.0 & 98.5/62.2 & 98.8/57.2 \\
carpet & 91.5/45.8 & \textbf{99.7}/67.0 & 96.3/62.5 & 92.3/49.1 & 99.1/47.7 & 99.4/46.0 & 98.7/65.0 & 94.6/19.6 & 98.2/54.0 & 99.0/\textbf{82.0} & 98.6/81.2 \\
grid & 89.0/\pzo 7.6 & 99.1/\textbf{87.8} & 99.5/53.2 & \textbf{99.6}/58.2 & 98.6/82.9 & 98.0/75.4 & 97.2/23.6 & 90.2/44.9 & 86.2/28.6 & 98.4/45.7 & 98.3/52.9 \\
hazelnut & 95.0/34.2 & 97.9/67.2 & 99.5/88.1 & 99.6/94.5 & 98.5/80.2 & 94.2/57.2 & 97.6/55.2 & 76.9/46.8 & 88.8/20.3 & 99.7/93.8 & \textbf{99.8}/\textbf{96.5} \\
leather & 98.2/26.7 & 99.2/\textbf{91.1} & 98.8/68.5 & 97.2/60.3 & 96.2/60.9 & 96.6/53.5 & 98.9/43.4 & 94.3/66.2 & 97.2/\pzo 5.1 & 99.7/69.7 & \textbf{99.8}/79.6 \\
metal nut & 81.7/30.6 & 98.8/78.2 & \textbf{98.7}/91.6 & 99.3/95.1 & 98.6/74.6 & 97.3/53.8 & 97.5/86.6 & 93.3/57.4 & 80.3/30.6 & 99.7/98.0 & \textbf{99.8}/\textbf{98.7} \\
pill & 90.1/23.1 & 98.9/60.3 & \textbf{97}.7/44.8 & 96.5/48.1 & 98.8/67.9 & 98.4/58.1 & \textbf{97.0}/75.9 & 98.9/79.9 & 79.6/22.1 & 99.5/91.3 & \textbf{99.8}/\textbf{97.0} \\
screw & 95.4/\pzo 5.9 & 98.8/45.7 & \textbf{99.7}/\textbf{72.9} & 99.1/62.0 & 98.7/61.4 & 99.1/51.8 & 98.7/34.2 & 66.5/21.1 & 51.0/\pzo 5.1 & 97.5/44.9 & 97.0/51.8 \\
tile & 78.6/26.7 & 98.0/86.7 & 99.4/96.4 & 99.2/96.3 & 98.6/92.6 & 97.4/78.2 & 94.9/56.0 & 88.7/63.9 & 91.0/54.4 & \textbf{99.6}/\textbf{96.5} & 99.2/93.9 \\
toothbrush & 95.6/20.0 & 99.1/56.9 & 97.3/49.2 & 97.5/38.9 & 98.4/61.7 & 99.0/63.1 & 97.6/37.1 & 96.3/52.4 & 74.5/\pzo 4.8 & \textbf{99.6}/\textbf{78.1} & 99.1/76.5 \\
transistor & 76.0/25.9 & 98.8/40.6 & 92.2/56.0 & 85.3/36.5 & 98.6/82.9 & \textbf{99.6}/50.3 & 91.8/66.7 & 55.2/\pzo 4.4 & 79.3/11.2 & 98.4/85.6 & 99.3/\textbf{92.6} \\
wood & 88.3/24.7 & 98.9/47.2 & 97.6/81.6 & 97.2/77.1 & 97.6/25.6 & \textbf{99.3}/39.1 & 95.7/54.3 & 93.1/47.9 & 82.9/21.0 & 97.8/82.6 & 98.9/\textbf{84.6} \\
zipper & 95.1/30.5 & 96.5/63.9 & 98.6/73.6 & 98.1/78.2 & 95.9/53.9 & \textbf{99.7}/52.7 & 98.5/63.1 & 92.4/53.1 & 96.8/42.3 & 98.8/77.6 & 99.4/\textbf{86.0} \\
\midrule
Average & 89.6/24.9 & 98.7/65.3 & 97.7/69.0 & 96.2/65.5 & 98.3/66.3 & 98.3/57.8 & 97.1/56.6 & 86.4/49.3 & 84.8/25.7 & 99.0/78.6 & \textbf{99.1}/\textbf{81.4} \\
\bottomrule
\end{tabular}}
\caption{\textbf{Comparison on pixel-level anomaly localization (AUROC/AP)} between the simple U-Net trained on our generated dataset and the existing anomaly detection methods \hut{ with their official codes or pre-trained models}.
}
\label{tab:pixel aupro and pixel-ap}
\end{table*} 


\begin{table}[t!]
\centering
\renewcommand{\arraystretch}{1.0}
\setlength\tabcolsep{12.0pt}
\resizebox{.9\linewidth}{!}{
\begin{tabular}{ccc|cccc}
\toprule
\multicolumn{3}{c}{Method} & \multicolumn{3}{c}{Metric} \\
SAE & Masked $\mathcal{L}$ & AAR  & AUROC & AP & $F_1$-max  \\
\midrule
 &  &  &  81.3 & 31.1 & 46.5\\
\checkmark &  &  &  90.3 & 51.2 & 60.7 \\
\checkmark & \checkmark &  & 95.0 &64.9&68.8\\
 & \checkmark& \checkmark & 95.5&67.5&68.9\\
\checkmark & \checkmark& \checkmark &  \textbf{99.1} & \textbf{81.4} & \textbf{76.3}\\
\bottomrule
\end{tabular}}
\caption{\textbf{Ablation study} on our spatial anomaly embedding (SAE), masked diffusion loss (Masked $\mathcal{L}$) and adaptive attention re-weighting mechanism (AAR).}
\label{tab:ablation study}
\end{table}

\noindent\textbf{Anomaly generation quality.}
We compare our model with DiffAug, CDC, Crop\&Paste, SDGAN, DefectGAN and DFMGAN on anomaly generation quality and diversity in Tab.~\ref{tab:comparison on kid and IC-LPIPS}. Since DRAEM and PRN crop random textures to imitate anomalies, we cannot compute IC-LPIPS for them. For each anomaly category, we allocate one-third of the anomaly data for training and generate 1000 anomaly images to compute IS and IC-LPIPS. It demonstrates that our model generates anomaly data with both the highest quality and diversity.

Moreover, we exhibit the generated anomalies in Fig.~\ref{fig:generation quality}. It can be seen that our model excels in producing high-quality authentic anomalies that accurately align with their corresponding masks. In contrast, 
CDC yields visually perplexing outcomes, particularly for structural anomaly categories like capsule-squeeze. SDGAN and DefectGAN yield poor outputs, frequently encountering difficulties in generating anomalies such as pill-crack. The state-of-the-art model DFMGAN sometimes struggles to produce authentic anomalies and fails to keep the alignment between the generated anomalies and masks, as shown in metal nut-bent. More results are presented in supplementary material.


\noindent\textbf{Anomaly generation for anomaly detection and localization.}
We compare the performance of our approach with existing anomaly generation methods in downstream anomaly detection and localization. Due to the inability of DiffAug and SDGAN to generate anomaly masks, we only compare our method with Crop\&Paste, DRAEM, PRN and DFMGAN. For each method, we generate 1000 images per anomaly category and train an U-Net~\cite{ronneberger2015unet} alongside normal samples for anomaly localization. The localization outcomes are aggregated using average pooling to derive confidence scores for image-level anomaly detection (the same as DREAM). We compute pixel-level metrics including AUROC, AP, $F_1$-max. The results, as presented in Tab.~\ref{tab:pixel level anomaly localization comparison}, illustrate that our model outperforms other anomaly generation models at most conditions.
Furthermore, we also evaluate image-level AUROC, AP, and $F_1$-max scores in Tab.~\ref{tab:image-level anomlay detection comparison}. It demonstrates our model has the best anomaly detection performance compared to other methods.
We also compare the qualitative results on anomaly localization in Fig.~\ref{fig: qualitative comparison}, which shows our superior performance in localizing the anomalies.



\noindent\textbf{Anomaly generation for anomaly classification.}
To further validate the generation quality of our model, we employ the generated anomalies to train a downstream anomaly classification model. Specifically, we adopt the experiment setting in DFMGAN, which trains a ResNet-34~\cite{he2016resnet} on the generated dataset and test the classification accuracy on the remaining shared test dataset. The comparison results are shown in Tab.~\ref{tab:comparison on anomaly classfication. }. It can be seen that our model outperforms all other models in almost all types of components and the average accuracy (\textbf{66.09\%}) surpasses that of the second-ranked DFMGAN (49.61\%) by a margin of \textbf{16.48\%}.


\subsection{Comparison with Anomaly Detection Models}

To further validate the efficacy of our model, we conduct a comparative experiment with the state-of-the-art anomaly detection methods CFLOW~\cite{gudovskiy2022cflow}, DRAEM~\cite{zavrtanik2021draem}, CFA~\cite{lee2022cfa}, RD4AD~\cite{deng2022rd4ad}, PatchCore~\cite{roth2022patchcore}, DevNet~\cite{pang2021devnet}, DRA~\cite{ding2022dra} and PRN~\cite{zhang2023prn}. We employ their official codes or pre-trained models and evaluate them on the same testing dataset that we use. It is worth noting that due to the absence of the open-source code for PRN, we utilize the data provided in its paper. The comparison results on pixel-level AUROC and AP are presented in Tab.~\ref{tab:pixel aupro and pixel-ap}. It can be seen that although our model is only a simple U-Net, with the help of our generated anomaly data, it has a good performance in anomaly localization with the highest AP of \textbf{81.4\%} and AUROC of \textbf{99.1\%}, indicating the profound significance of our generated data for downstream anomaly inspection tasks.

\subsection{Ablation Study}

\hut{We evaluate the effectiveness of our components: spatial anomaly embedding (SAE), masked diffusion loss (Masked $\mathcal{L}$) and adaptive attention re-weighting mechanism (AAR). Not that the models without SAE employ only an anomaly embedding trained by textual inversion. We train 5 models: \textbf{\textit{1)}} with none of these components; \textbf{\textit{2)}} only SAE; \textbf{\textit{3)}} SAE + masked $\mathcal{L}$; \textbf{\textit{4)}} masked $\mathcal{L}$ + AAR and \textbf{\textit{5)}} the full model (ours). We employ these models to generate 1000 anomalous image-mask pairs and train an U-Net for anomaly localization. We compare the pixel-level localization results in Tab.~\ref{tab:ablation study}. It demonstrates that the omission of any of the proposed modules leads to a noticeable decline in the model's performance on anomaly localization, 
\yr{which validates}
the efficacy of the \yr{proposed} modules. For more experiments, please refer to the supplementary material.
}

\section{Conclusion}
In this paper, we propose \textit{Anomalydiffusion}, a novel anomaly generation model which generates anomalous image-mask pairs. We disentangle anomaly information into anomaly appearance and location information represented by anomaly embedding and spatial embedding in the textual space of LDM. 
Moreover, we also introduce an adaptive attention re-weighting mechanism, which helps our model focus more on the areas with less noticeable generated anomalies, thus improving the alignment between the generated anomalies and masks. Extensive experiments show that our model outperforms the existing anomaly generation methods and our generated anomaly data 
\yr{effectively improves the performance of}
the downstream anomaly inspection tasks. In future work, we would explore the application of a more potent diffusion model to enhance the resolution of the generated anomalies, which could further improve the performance.

\section*{Acknowledgments}
This work was supported by National Natural Science Foundation of China (62302297, 72192821, 62272447), Young Elite Scientists Sponsorship Program by CAST (2022QNRC001), Shanghai Sailing Program (22YF1420300), Beijing Natural Science Foundation (L222117), the Fundamental Research Funds for the Central Universities (YG2023QNB17), Shanghai Municipal Science and Technology Major Project (2021SHZDZX0102), Shanghai Science and Technology Commission  (21511101200), CCF-Tencent Open Research Fund (RAGR20220121).

\bibliography{aaai24}

\newpage
\appendix

\section{Overview}
This supplementary material consists of:
\begin{itemize}
    \item Details of the data augmentation method (Sec.~\ref{sec:data augmentation}).
    \item More implementation details (Sec.~\ref{sec:Implementation details}).
    \item More ablation studies (Sec.~\ref{sec:more ablation study}).
    \item Comparison between our Spatial Anomaly Embedding and Prompt-to-Prompt (Sec.~\ref{sec:comparsion with p2p}).
    \item More qualitative comparison results with the anomaly generation methods (Sec.~\ref{sec:more qualitative comparison}).
    \item More quantitative comparison results with the anomaly generation methods (Sec.~\ref{sec:more quantitative compairson}).
    
\end{itemize}

\section{Data Augmentation}
\label{sec:data augmentation}
Due to the limited number of samples for each anomaly category, typically less than 10 images are available for training. This constraint makes it challenging for our anomaly embedding to completely eliminate spatial information, as it still tends to generate anomalies at the positions observed in the training images. Additionally, when the training data for the spatial encoder is scarce, the model becomes susceptible to overfitting, making it difficult to accurately generate anomalies at the correct positions.

To address these issues, we employ a data augmentation approach during training. For paired image-mask data, we perform random cropping, translation, and rotation on both the image and its corresponding mask. By recording the maximum and minimum coordinates of the anomaly region in the image, we ensure that the anomaly remains within the image during data augmentation. This data augmentation process effectively disrupts the spatial information within the training data, causing the anomaly embedding to lose its focus on recording positions and instead concentrate solely on the anomaly appearance. Simultaneously, the spatial encoder benefits from having enough  augmented data for training, boosting its ability in position encoding.

\section{Implementation Details}
\label{sec:Implementation details}

\subsection{Training Details}
\textbf{Training spatial anomaly embedding.}
For each anomaly type, an anomaly embedding $e_a$ is assigned, while a shared spatial encoder $E$ is employed across all anomaly categories. Each anomaly embedding $e_a$ is composed of 8 tokens, and the spatial embedding $e_s$ comprises 4 tokens. The batch size is set at 4, and the learning rate is 0.005. During each training iteration, we randomly sample 4 anomalous image-mask pairs from all the anomaly categories.  We train all the anomaly embedding and spatial encoder at the same time for 300K iterations \hut{in 3 days on an NVIDIA GeForce RTX 3090 24GB GPU}.

\noindent\textbf{Training mask embedding.} For each anomaly type, we assign a mask embedding $e_m$ for it. To enhance the diversity of the generated masks, each mask embedding consists of 2 tokens, preventing it from overfitting. Furthermore, with a batch size of 4 and a learning rate of 0.005, each mask embedding is trained for 30Kiterations.

\noindent\textbf{Mask Generation.} 
With the trained mask embedding, we input it as a text condition to guide the generation process of latent diffusion model~\cite{rombach2022ldm}. Specifically, we employ the classifier-free guidance~\cite{ho2022classifierfree} to generate masks:

\begin{equation}
\hat{\epsilon}_\theta\left(x_t \mid e_m\right)=\epsilon_\theta\left(x_t\right)+s \cdot\left(\epsilon_\theta\left(x_t, e_m\right)-\epsilon_\theta\left(x_t\right)\right),
\end{equation}
where s is set 5 (the same as Textual Inversion).

\subsection{Metrics}

\hut{In the quantitative experiments, we employ the following metrics to measure the model performance.
\begin{itemize}
    \item \textbf{Inception Score(IS)} quantifies the quality and diversity of generated images by computing the exponential of the negative of the KL divergence between the marginal distribution of generated images and the conditional distribution of class labels predicted by an Inception model. A higher IS score represents a better generation quality and diversity.
    \item \textbf{Intra-cluster Pairwise LPIPS Distance (IC-LPIPIS)}~\cite{ojha2021few} clusters the generated images into $k$ groups based on LPIPS distance to $k$ target samples, and then compute the average mean LPIPS distances to corresponding target samples within each cluster. A higher IC-LPIPS indicates a better generation diversity.
    \item \textbf{Area Under the Receiver Operating Characteristic (AUROC)} measures the performance of a binary classification model by evaluating its ability to distinguish between true positive and false positive rates across different probability thresholds. A higher AUROC means better anomaly detection and localization performance.
    \item \textbf{Average Precision (AP, which is also known as PR-AUC)} assesses the precision-recall curve for a classification model, calculating the average precision of the model across different recall levels, providing a summary of its overall performance. A higher AP means better anomaly detection and localization results.
    \item \textbf{$\mathbf{F_1}$-max} is a variant of the $F_1$ score that maximizes both precision and recall by selecting the threshold that yields the highest F1 score when evaluating a binary classification model. A higher $\mathbf{F_1}$-max represents better anomaly detection and localization results 
\end{itemize}}

\section{More Ablation Studies}

\label{sec:more ablation study}
\subsection{Ablation on Spatial Anomaly Embedding}

\begin{figure}[t]
\centering
\includegraphics[width=0.48\textwidth]{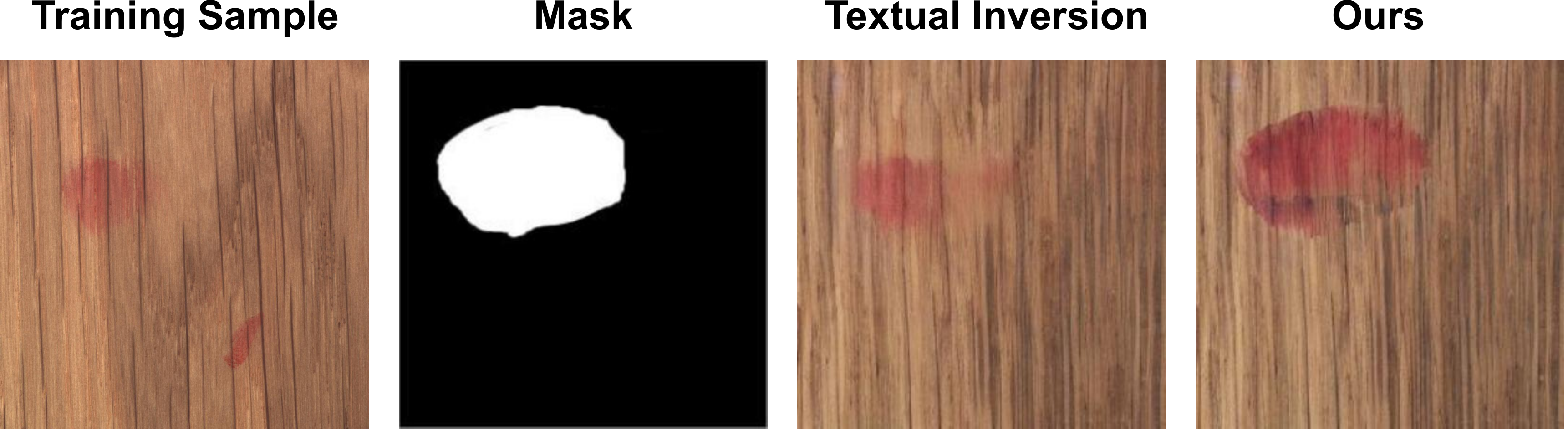}
\caption{\textbf{Comparison results between Textual Inversion (Anomaly Embedding only) and our model (Spatial Anomaly Embedding).} The generated result of Textual Inversion tends to generate anomalies at the location as the same as the training sample.}
\label{fig: compare with textual inversion}
\end{figure}

We aim to seek a text embedding that guides the latent diffusion model in generating anomalies within a given anomaly mask.  However, textual inversion tends to capture the location of anomalies along with the anomaly type information, which results in the generated anomalies only distributed in specific locations.
Therefore, we propose spatial anomaly embedding $e$, consisting of an anomaly embedding $e_a$ (for appearance) and a spatial embedding $e_s$ (for location), which disentangles the spatial information from anomaly appearance. To further validate this theory, we directly employ text inversion to train an anomaly embedding and use the trained embedding to generate anomalies with a given mask by blended latent diffusion (the same as our generation process). The generated results are shown in Fig.~\ref{fig: compare with textual inversion}. It can be seen that the generated result by Textual Inversion tends to generate anomalies at the location the same as the training sample, which limits its application in anomaly generation where anomalies can be located at arbitrary positions.

\subsection{Ablation on the rate of anomalies}

\hut{We conduct additional experiments with the rate of anomalies as 10\%, 20\%, 30\% (\textbf{Ours}), 40\%, and 50\% and test the performance on anomaly localization measured by AUROC and AP. The performance of anomaly localization is shown in Tab.~\ref{tab:ablation on the anomaly rate}.
It can be seen that AP decreases quickly when the anomaly rate falls below 30\%. This is attributed to the limited availability of training data for most categories, often comprising only 1-2 instances, making it challenging for the model to capture the anomaly information. 
Conversely, when the rate exceeds 30\%, the model performance is similar. This indicates that our model can effectively learn sufficient anomaly information without a heavy reliance on an abundance of training samples. }
\begin{table}[t]
\centering
\renewcommand{\arraystretch}{1.0}
\setlength\tabcolsep{6.0pt}
\resizebox{1.0\linewidth}{!}{
\begin{tabular}{c|ccccc}
\toprule
Anomaly Rate&10\%&20\%&\textbf{30\% (Ours)}&40\%&50\%\\ \hline
AUROC $\uparrow$ & 96.2&98.1&\textbf{99.1}&99.0&98.7\\
AP $\uparrow$ &64.2&75.5&\textbf{81.4}&81.1&80.0\\
\bottomrule
\end{tabular}}
\caption{\hut{\textbf{ablation study on the rate of anomalies.}}}
\label{tab:ablation on the anomaly rate}
\end{table}

\subsection{Ablation on the hyperparameters}
\hut{We conduct ablation studies on the length of the anomaly embedding $l_a$ and spatial embedding $l_s$ in Tab.~\ref{tab:ablation on the hyperparameters}. Specifically, we train models with different $l_a$ and $l_s$ and then employ their generated data to train a UNet to localize the anomalies. It can be seen that when increasing $l_s$ and $l_a$, the total parameter number of the model rises, but the final performance in the downstream anomaly localization task is similar, which demonstrates that our model is not snesitive to the hyperparameters. }

\begin{table}[t]

\centering
\renewcommand{\arraystretch}{1.1}
\setlength\tabcolsep{6.0pt}
\resizebox{1.\linewidth}{!}{
\begin{tabular}{l|cccc}
\toprule
 \makecell[c]{Model} & AUROC $\uparrow$ & AP $\uparrow$ & $F_1$-max $\uparrow$& PRO $\uparrow$\\ \hline
 $L_s=4,l_a=8$ (\textbf{Ours})&\textbf{99.1}&\textbf{81.4}&\textbf{76.3}&\textbf{94.0}\\
 $L_s=4,l_a=16$&98.8&80.6&75.1&93.2\\
 $L_s=8,l_a=8$&99.0&80.9&75.8&93.5\\
 $L_s=8,l_a=16$&\textbf{99.1}&81.2&75.9&93.8\\
\bottomrule
\end{tabular}}
\caption{\hut{\textbf{Ablation study on the anomaly embedding $l_a$ and spatial embedding $l_s$}}}
\label{tab:ablation on the hyperparameters}
\end{table}

\subsection{Ablation on SAE and AAR}

\begin{figure}[t]
\centering
\includegraphics[width=0.48\textwidth]{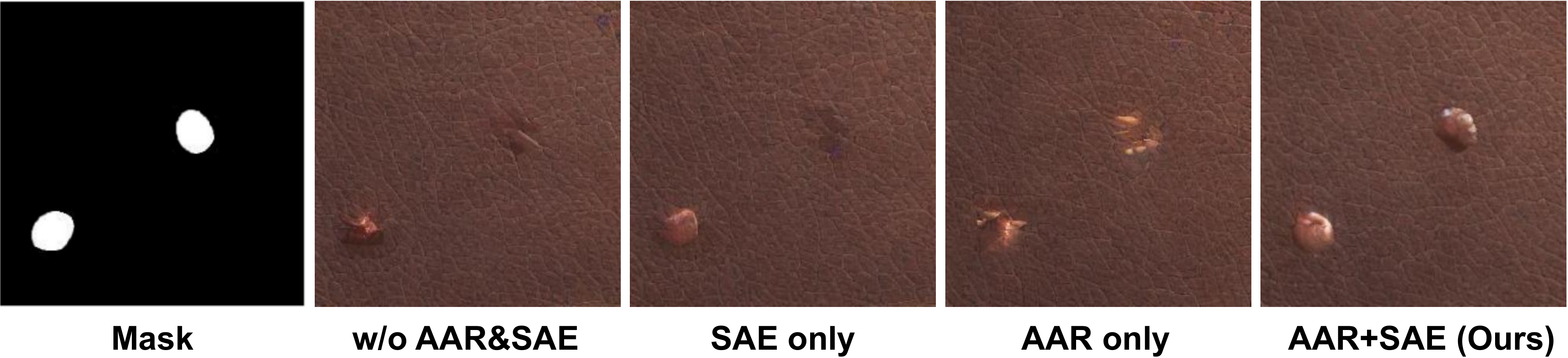}
\caption{\hut{\textbf{Ablation study on SAE and AAR.}}}
\label{fig:ablation on SAE-AAR}
\end{figure}

\hut{ We conduct more ablation studies on the effectiveness of our spatial anomaly embedding (SAE) and adaptive attention re-weighting mechanism (AAR) by adding SAE and AAR separately.
We compare our model with 3 models: \textbf{\textit{1)}} w/o AAR\&SAE, \textbf{\textit{2)}} AAR only, and \textbf{\textit{3)}} SAE only in generating glue anomalies to the leather in the Fig.~\ref{fig:ablation on SAE-AAR}.
It shows that the model without AAR\&SAE cannot generate authentic anomalies or fill anomaly mask. While adding SAE improves anomaly authenticity, it doesn't fill the mask. Moreover, incorporating AAR fills the mask but sacrifices authenticity. In contrast, our model (SAE + AAR) effectively generates authentic anomalies filling the mask.}

\section{Comparison with Prompt-to-Prompt}
\label{sec:comparsion with p2p}
Prompt-to-Prompt~\cite{hertz2022prompt} proposed a method that allows modifying generated images by altering corresponding text descriptions. For instance, when transforming "a cat sits on the street" to "a dog sits on the street," Prompt-to-Prompt replaces the cross-attention map of "dog" with that of "cat", which transforms the cat in the original image into a dog while maintaining nearly unchanged content in other regions, achieving controlled image generation with specific positions. However, Prompt-to-Prompt requires a text corresponding to the original image for generation, which is unavailable in anomaly generation.

It seems that Prompt-to-prompt presents a potential solution for controlling generation positions through the manipulation of cross-attention maps. A direct solution is to resize the mask $m$ to match and substitute the cross-attention map $m_c$ of anomaly embedding, thus controlling the generation location. However, even though the new cross-attention map $m'_c$ seemingly dictates anomaly location, it could conflict with the values $V$ in the cross-attention module. Since $V$ is designed for the original cross-attention map $m_c$, the semantic information of $V$ in the newly enforced mask $m'_c$ might not align with the semantics in original mask $m_c$, consequently leading to a unstable generated results.

\begin{figure}[t]
\centering
\includegraphics[width=0.48\textwidth]{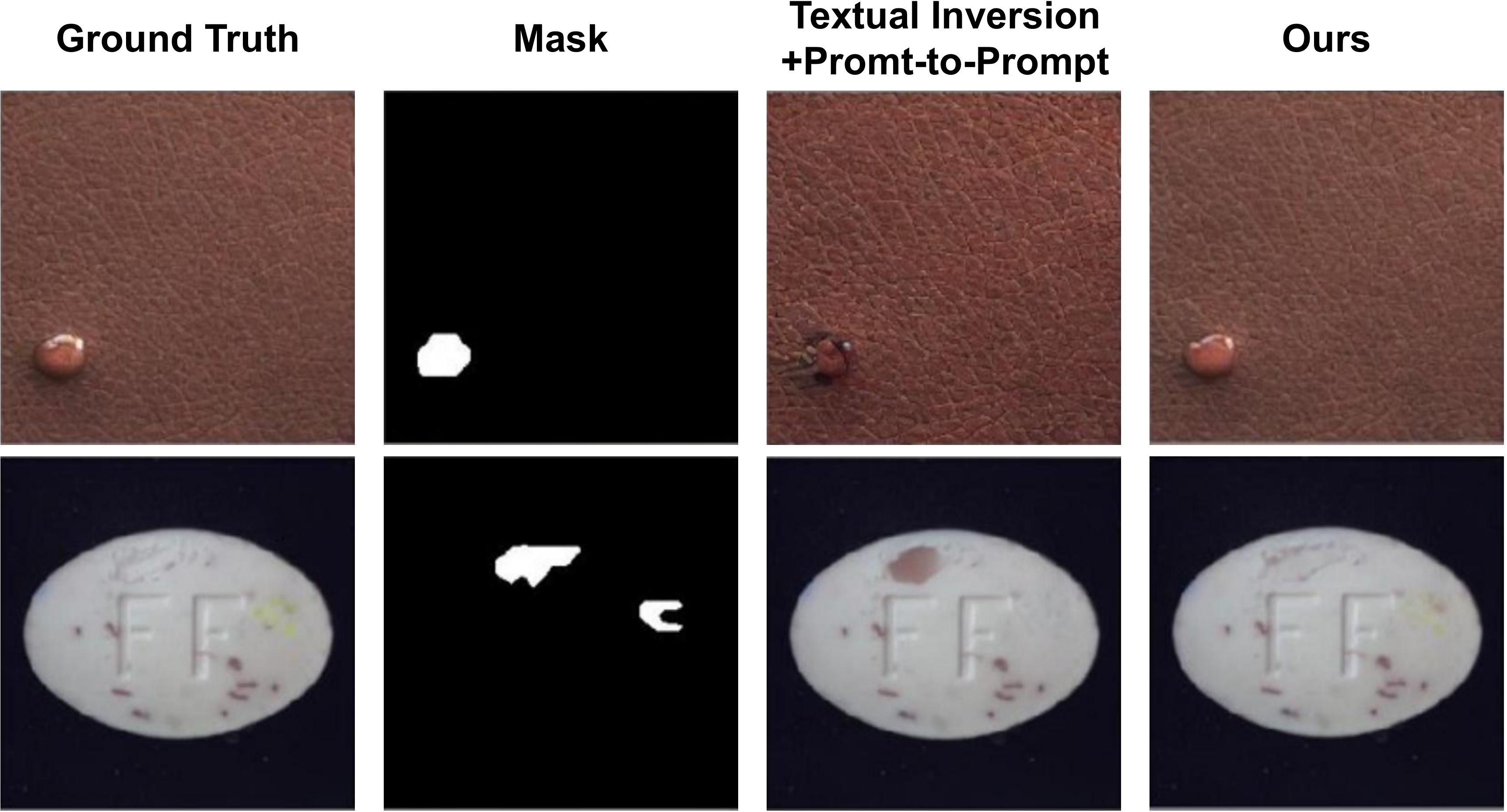}
\caption{\textbf{Comparison results between Textual Inversion + Prompt-to-Prompt and our model (Spatial Anomaly Embedding) in anomaly generation.}}
\label{fig: compare with p2p}

\end{figure}

To verify it, we conduct reconstruction experiments on real anomalies, comparing the results of Textual Inversion + Prompt-to-Prompt with our approach (Spatial Anomaly Embedding). Specifically, we sample a real anomaly image $I$ as ground truth and mask out the anomaly parts for generation. The results are shown in Fig.~\ref{fig: compare with p2p}. Textual Inversion + Prompt-to-Prompt can not generate anomalies as authentic as ours. And its generated anomalies are quite different from the ground truth, indicating that replacing the cross-attention map directly cannot generate satisfying anomalies. Moreover, we also conduct a quantitative experiment, where we generate anomalous image-mask pairs to support the downstream anomaly localization task. We follow the experiment settings in the main paper, in which we train an U-Net on the  generated data and compare the localization accuracy. The results are recorded in Tab.~\ref{tab:comparison with p2p}. Our model outperforms Textual Inversion + Prompt-to-Prompt significantly. 

\begin{table}[t]
\renewcommand{\arraystretch}{1.1}
\setlength\tabcolsep{6.0pt}
\resizebox{1.\linewidth}{!}{
\begin{tabular}{c|cccc}
\toprule
\multirow{2}{*}{Method} & \multicolumn{4}{c}{Metric} \\
  & AUROC & AP & $F_1$-max & PRO \\
\midrule
\makecell[c]{Textual Inversion+\\Prompt-to-Prompt} & 91.2 & 55.1 & 64.4 & 73.5 \\
Ours& \textbf{99.1} & \textbf{81.4} & \textbf{76.4} & \textbf{94.0}\\
\bottomrule
\end{tabular}}
\caption{\textbf{Comparison with Textual Inversion + Prompt-to-Prompt on anomaly localization.}}
\label{tab:comparison with p2p}
\end{table}

\section{More qualitative experiments}
\label{sec:more qualitative comparison}
\begin{figure*}[t]
\centering
\includegraphics[width=0.98\textwidth]{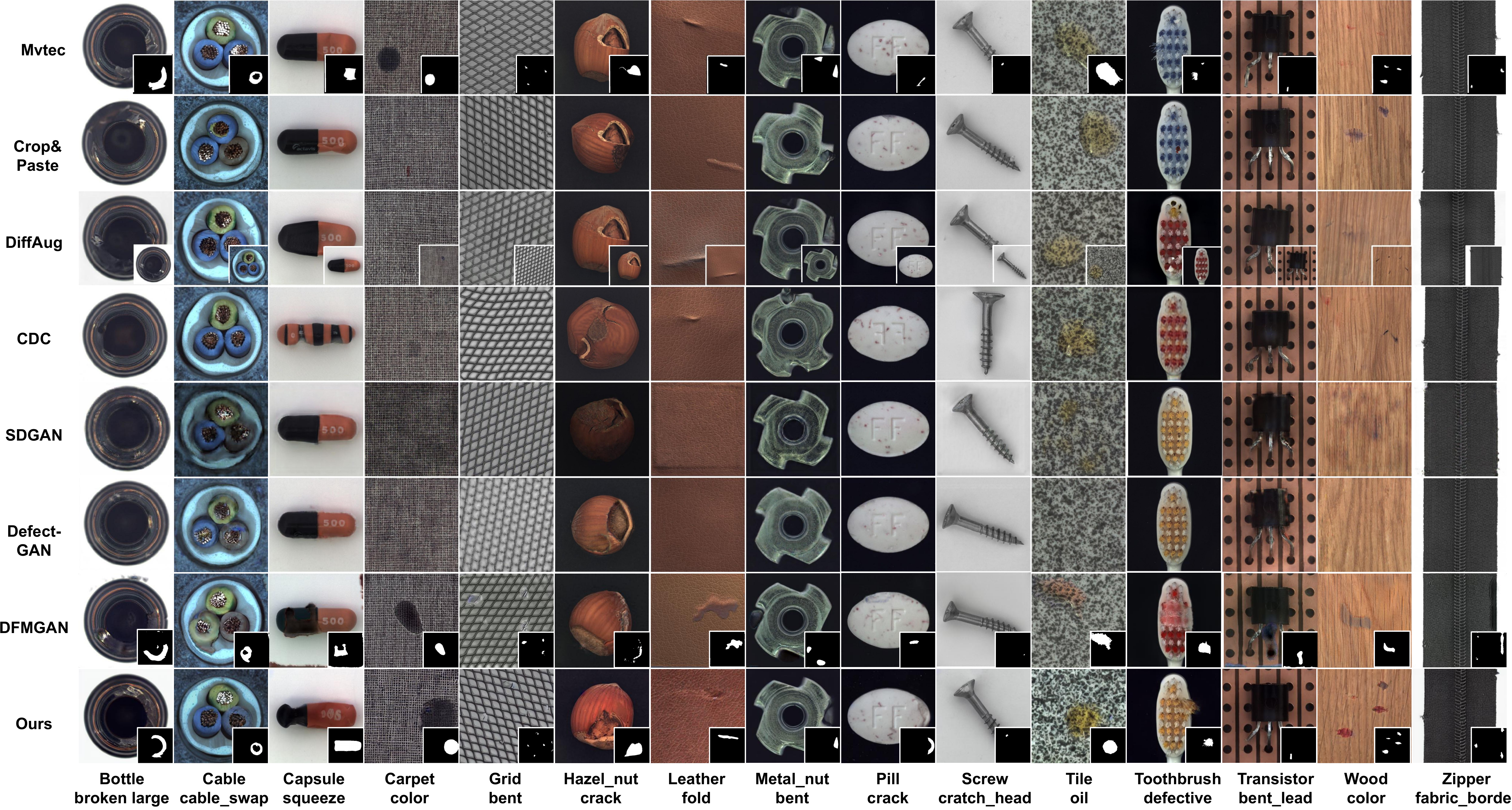}
\caption{\textbf{Qualitative comparison on the anomaly generation quality.} Note that the generated anomlalies by DiffAug is the same as the training samples (images in lower-right corner)}
\label{fig: suppl_qualitative comparison}

\end{figure*}

 We give a more comprehensive comparison with the existing anomaly generation methods DiffAug~\cite{zhao2020diffaug}, CDC~\cite{ojha2021few}, Crop\&Paste, SDGAN~\cite{niu2020sdgan}, Defect-GAN~\cite{zhang2021defectgan} and DFMGAN~\cite{duan2023DFMGAN}. We exhibit the generation results of all the anomaly generation methods across all components in Fig.~\ref{fig: suppl_qualitative comparison}.
Our model demonstrates remarkable proficiency in generating high-quality, authentic anomalies that are precisely aligned with the corresponding masks. In contrast, Crop\&Paste exhibits limited diversity in generating various anomaly types. DiffAug displays evident overfitting tendencies towards the training samples (the image in the lower-right corner). CDC yields visually perplexing results, particularly for structural anomaly categories like capsule-squeeze. SDGAN and DefectGAN yield poor outputs, frequently encountering challenges in generating anomalies such as pill-crack. The state-of-the-art model DFMGAN occasionally struggles to create authentic anomalies and fails in maintaining alignment between the generated anomalies and masks, as observed in the case of metal nut-bent. In comparison, our model generates anomalies with the highest diversity and authenticity, and the generated anomalies align with the masks accurately, which can effectively support the downstream anomaly inspection tasks.

\section{More Quantitative experiments}
\label{sec:more quantitative compairson}
{\bf More comparison with the anomaly generation models.} In this section, we provide supplementary experiments to complement those presented in the main paper. Specifically, in addition to the methods covered in the main paper, we include Crop\&Paste~\cite{lin2021croppaste} for comparison and we additionally introduce the Per Region Overlap (PRO) metric to provide a more comprehensive evaluation on anomaly localization. The experiment settings are the same as that in the main paper, where we train an U-net on the generated anomaly data. The pixel-level anomaly localization results are shown in Tab.~\ref{tab:suppl_pixel level anomaly localization comparison} and the image-level anomaly detection results are shown in Tab.~\ref{tab:suppl_image-level anomlay detection comparison}. The quantitative results demonstrate that our model outperforms all the other anomaly generation methods in terms of both anomaly localization and detection, indicating our good anomaly generation quality and diversity.

\hut{{\bf More comparison with the anomaly localization models.} In this section,  we further compare the anomaly detection methods with $F_1$-max score on anomaly localization. The results are shown in Table \ref{tab:comparison on anomaly localization with f_1 max. }. It can be seen that our model achieves the best performance in anomaly localization with $F_1$-max.}

\begin{table*}[t]
\centering
\renewcommand{\arraystretch}{1.0}
\setlength\tabcolsep{2.0pt}
\resizebox{1.0\linewidth}{!}{
\begin{tabular}{c|cccc|cccc|cccc|cccc|cccc}
\toprule
\multirow{2}{*}{Category} & \multicolumn{4}{c}{Crop\&Paste} & \multicolumn{4}{c}{DRAEM} & \multicolumn{4}{c}{PRN} & \multicolumn{4}{c}{DFMGAN} & \multicolumn{4}{c}{Ours} \\
 & AUC & AP & $F_1$-max & PRO & AUC & AP & $F_1$-max & PRO & AUC & AP & $F_1$-max & PRO & AUC & AP & $F_1$-max & PRO & AUC & AP & $F_1$-max & PRO \\
\midrule
bottle & 94.5 & 67.4 & 63.5 & 77.8 & 96.7 & 80.2 & 74.0 & 91.2 & 97.5 & 76.4 & 71.3 & 88.5 & \underline{98.9} & \underline{90.2} & \underline{83.9} & \underline{91.7} & \textbf{99.4} & \textbf{94.1} & \textbf{87.3} & \textbf{94.3} \\
cable & 96.0 & 75.3 & 69.3 & \underline{87.1} & 80.3 & 21.8 & 28.3 & 58.2 & 94.5 & 64.4 & 61.0 & 79.7 & \underline{97.2} & \underline{81.0} & \underline{75.4} & 84.9 & \textbf{99.2} & \textbf{90.8} & \textbf{83.5} & \textbf{95.0} \\
capsule & 95.3 & \underline{49.2} & \underline{51.1} & 89.5 & 76.2 & 25.5 & 32.1 & 81.1 & \underline{95.6} & 45.7 & 47.9 & \underline{89.7} & 79.2 & 26.0 & 35.0 & 66.1 & \textbf{98.8} & \textbf{57.2} & \textbf{59.8} & \textbf{95.4} \\
carpet & 83.7 & 36.6 & 39.7 & 62.9 & 92.6 & 43.0 & 41.9 & 80.0 & \underline{96.4} & \underline{69.6} & \underline{65.6} & \underline{90.6} & \underline{90.6} & 33.4 & 38.1 & 76.5 & \textbf{98.6} & \textbf{81.2} & \textbf{74.6} & \textbf{91.6} \\
grid & 84.7 & 13.1 & 22.4 & 70.2 & \textbf{99.1} & \textbf{59.3} & \underline{58.7} & \textbf{95.8} & \underline{98.9} & \underline{58.6} & \textbf{58.9} & \textbf{95.8} & 75.2 & 14.3 & 20.5 & 52.3 & 98.3 & 52.9 & 54.6 & \underline{92.3} \\
hazelnut & 88.5 & 38.0 & 42.8 & 74.1 & 98.8 & 73.6 & 68.5 & 95.9 & 98.0 & 73.9 & 68.2 & 92.7 & \underline{99.7} & \underline{95.2} & \underline{89.5} & \underline{96.4} & \textbf{99.8} & \textbf{96.5} & \textbf{90.6} & \textbf{97.1} \\
leather & \underline{97.5} & \underline{76.0} & \underline{70.8} & 95.7 & 98.5 & 67.6 & 65.0 & 96.7 & \underline{99.4} & 58.1 & 54.0 & \underline{97.5} & 98.5 & 68.7 & 66.7 & 96.0 & \textbf{99.8} & \textbf{79.6} & \textbf{71.0} & \textbf{98.2} \\
metal nut & 96.3 & 84.2 & 74.0 & 67.2 & 96.9 & 84.2 & 74.5 & \underline{90.4} & 97.9 & 93.0 & 87.1 & 85.0 & \underline{99.3} & \underline{98.1} & \textbf{94.5} & 88.0 & \textbf{99.8} & \textbf{98.7} & \underline{94.0} & \textbf{94.8} \\
pill & 81.5 & 17.8 & 24.3 & 57.4 & 95.8 & 45.3 & 53.0 & 83.7 & \underline{98.3} & 55.5 & \underline{72.6} & \underline{88.2} & 81.2 & \underline{67.8} & \underline{72.6} & 56.5 & \textbf{99.8} & \textbf{97.0} & \textbf{90.8} & \textbf{97.3} \\
screw & 93.4 & 31.2 & 36.0 & \textbf{83.9} & 91.0 & 30.1 & 35.7 & 78.1 & \underline{94.0} & \underline{47.7} & \underline{49.8} & \underline{83.8} & 58.8 & 2.2 & 5.3 & 41.8 & \textbf{97.0} & \textbf{51.8} & \textbf{50.9} & 80.3 \\
tile & 94.0 & 79.3 & 74.5 & 79.2 & 98.5 & 93.2 & \underline{87.8} & 95.3 & 98.5 & 91.8 & 84.4 & 91.3 & \textbf{99.5} & \textbf{97.1} & \textbf{91.6} & \textbf{97.5} & \underline{99.2} & \underline{93.9} & 86.2 & \underline{96.1} \\
toothbrush & 89.3 & 30.9 & 34.6 & 66.6 & 93.8 & 29.5 & 28.4 & 75.1 & 96.1 & 46.4 & 46.2 & \underline{83.1} & \underline{96.4} & \underline{75.9} & \underline{72.6} & 74.3 & \textbf{99.2} & \textbf{76.5} & \textbf{73.4} & \textbf{91.4} \\
transistor & 85.9 & 52.5 & 52.1 & 64.5 & 76.5 & 31.7 & 24.2 & 54.3 & 94.9 & 68.6 & 68.4 & \underline{70.0} & \underline{96.2} & \underline{81.2} & \underline{77.0} & 65.5 & \textbf{99.3} & \textbf{92.6} & \textbf{85.7} & \textbf{96.2} \\
wood & 84.0 & 45.7 & 48.0 & 57.9 & \underline{98.8} & \textbf{87.8} & \textbf{80.9} & \textbf{94.7} & 96.2 & 74.2 & 67.4 & 82.1 & 95.3 & 70.7 & 65.8 & 89.9 & \textbf{98.9} & \underline{84.6} & \underline{74.5} & \underline{94.3} \\
zipper & 94.8 & 47.6 & 51.4 & 83.4 & 93.4 & 65.4 & 64.7 & 84.6 & \underline{98.4} & \underline{79.0} & \underline{73.7} & \underline{93.7} & 92.9 & 65.6 & 64.9 & 83.0 & \textbf{99.4} & \textbf{86.0} & \textbf{79.2} & \textbf{96.3} \\
\midrule
Average & 90.4 & 48.4 & 49.4 & 74.3 & 92.2 & 54.1 & 53.1 & 83.1 & \underline{96.9} & \underline{66.2} & \underline{64.7} & \underline{87.4} & 90.0 & 62.7 & 62.1 & 76.3 & \textbf{99.1} & \textbf{81.4} & \textbf{76.3} & \textbf{94.0} \\
\bottomrule

\end{tabular}}
\caption{\textbf{Comparison on the pixel-level anomaly localization} with AUC, AP, $F_1$-max and PRO metrics by training an U-Net on the generated datasets produced by Crop\&Paste, DRAEM, PRN, DFMGAN and our model. \textbf{Bold} and \underline{underline} represent optimal and sub-optimal results, respectively.}
\label{tab:suppl_pixel level anomaly localization comparison}
\end{table*}

\begin{table*}[t]
\renewcommand{\arraystretch}{1.0}
\setlength\tabcolsep{2.0pt}
\resizebox{1.\linewidth}{!}{
\begin{tabular}{c|ccc|ccc|ccc|ccc|ccc}
\toprule
\multirow{2}{*}{Category} & \multicolumn{3}{c}{Crop\&Paste} & \multicolumn{3}{c}{DRAEM} & \multicolumn{3}{c}{PRN} & \multicolumn{3}{c}{DFMGAN} & \multicolumn{3}{c}{Ours} \\
 & AUC & AP & $F_1$-max & AUC & AP & $F_1$-max & AUC & AP & $F_1$-max & AUC & AP & $F_1$-max & AUC & AP & $F_1$-max \\
 \midrule
bottle&85.4&95.1&90.9&\underline{99.3}&\underline{99.8}&\textbf{98.9}&94.9&98.4&94.1&\underline{99.3}&\underline{99.8}&\underline{97.7}&\textbf{99.8}&\textbf{99.9}&\textbf{98.9}\\
cable&93.3&96.1&91.6&72.1&83.2&79.2&86.3&92.0&84.0&\underline{95.9}&\underline{97.8}&\underline{93.8}&\textbf{100}&\textbf{100}&\textbf{100}\\
capsule&77.1&94.1&90.4&\underline{93.2}&\underline{98.7}&94.0&84.9&95.8&94.3&92.8&98.5&\underline{94.5}&\textbf{99.7}&\textbf{99.9}&\textbf{98.7}\\
carpet&57.7&84.3&87.3&\underline{95.3}&\underline{98.7}&\underline{93.4}&92.6&97.8&92.1&67.9&87.9&87.3&\textbf{96.7}&\textbf{98.8}&\textbf{94.3}\\
grid&83.0&94.1&87.6&\textbf{99.8}&\textbf{99.9}&\textbf{98.8}&96.6&98.9&95.0&73.0&90.4&85.4&\underline{98.4}&\underline{99.5}&\underline{98.7}\\
hazelnut&68.8&85.0&78.0&\textbf{100}&\textbf{100}&\textbf{100}&93.6&96.0&94.1&\underline{99.9}&\textbf{100}&\underline{99.0}&99.8&\underline{99.9}&98.9\\
leather&91.9&97.5&90.9&\textbf{100}&\textbf{100}&\textbf{100}&99.1&\underline{99.7}&97.6&\underline{99.9}&\textbf{100}&\underline{99.2}&\textbf{100}&\textbf{100}&\textbf{100}\\
metal nut&92.2&98.1&93.3&97.8&99.6&97.6&97.8&99.5&96.9&\underline{99.3}&\underline{99.8}&\underline{99.2}&\textbf{100}&\textbf{100}&\textbf{100}\\
pill&51.7&87.1&91.4&\underline{94.4}&\underline{98.9}&\underline{95.8}&88.8&97.8&93.2&68.7&91.7&91.4&\textbf{98}&\textbf{99.6}&\textbf{97}\\
screw&59.3&81.9&86.0&\underline{88.5}&\underline{96.3}&\underline{89.3}&84.1&94.7&87.2&22.3&64.7&85.3&\textbf{96.8}&\textbf{97.9}&\textbf{95.5}\\
tile&73.8&91.1&83.8&\textbf{100}&\textbf{100}&\textbf{100}&\underline{91.1}&\underline{96.9}&\underline{89.3}&\textbf{100}&\textbf{100}&\textbf{100}&\textbf{100}&\textbf{100}&\textbf{100}\\
toothbrush&81.2&91.0&88.9&\underline{99.4}&\underline{99.8}&\underline{97.6}&\textbf{100}&\textbf{100}&\textbf{100}&\textbf{100}&\textbf{100}&\textbf{100}&\textbf{100}&\textbf{100}&\textbf{100}\\
transistor&85.9&81.8&80.0&79.6&80.5&71.4&88.2&88.9&84.0&\underline{90.8}&\underline{92.5}&\underline{88.9}&\textbf{100}&\textbf{100}&\textbf{100}\\
wood&49.5&81.2&86.6&\textbf{100}&\textbf{100}&\textbf{100}&77.5&92.7&86.7&\underline{98.4}&\underline{99.4}&\underline{98.8}&\underline{98.4}&\underline{99.4}&\underline{98.8}\\
zipper&59.4&82.8&88.9&\textbf{100}&\textbf{100}&\textbf{100}&98.7&99.7&97.6&99.7&\underline{99.9}&\underline{99.4}&\underline{99.9}&\textbf{100}&\underline{99.4}\\
\midrule
Average&74.0&89.4&87.7&\underline{94.6}&\underline{97.0}&94.4&91.6&96.6&92.4&87.2&94.8&\underline{94.7}&\textbf{99.2}&\textbf{99.7}&\textbf{98.7}\\
\bottomrule

\end{tabular}}
\caption{\textbf{Comparison on the image-level anomaly detection} with AUC, AP and $F_1$-max metrics by training an U-Net on the generated datasets produced by Crop\&Paste, DRAEM, PRN, DFMGAN and our model.}
\label{tab:suppl_image-level anomlay detection comparison}
\end{table*}



\begin{table*}[t!]
\renewcommand{\arraystretch}{1.0}
\setlength\tabcolsep{5.0pt}
\resizebox{1.\linewidth}{!}{
\begin{tabular}{c|cccccccccc}
\toprule
$F_1$-max & KDAD & CFLOW & DREAM & SSPCAB & CFA & RD4AD & PatchCore & DevNet & DRA & Ours \\
\midrule
bottle & 50.9 & 9.5 & \underline{83.0} & 80.3 & 75.9 & 82.1 & 78.6 & 64.6 & 53.5 & \textbf{87.3} \\
cable & 18.3 & 10.5 & 58.5 & 51.2 & \underline{76.3} & 65.2 & 68.5 & 54.9 & 55.3 & \textbf{83.5} \\
capsule & 15.1 & 6.9 & 48.9 & 49.5 & 57.0 & \underline{60.4} & 56.7 & 38.7 & 47.7 & \textbf{60.8} \\
carpet & 54.2 & 3.5 & 60.0 & 47.1 & 48.3 & \underline{67.8} & 67.9 & 52.3 & 42.3 & \textbf{74.6} \\
grid & 10.9 & 3.2 & 56.3 & \underline{58.4} & 32.2 & \textbf{59.9} & 49.1 & 42.9 & 50.1 & 54.6 \\
hazelnut & 37.5 & 3.9 & 80.6 & \underline{88.9} & 61.4 & 70.0 & 68.1 & 22.4 & 47.2 & \textbf{90.6} \\
leather & 30.4 & 3.9 & 63.2 & 58.1 & 53.8 & \underline{67.2} & 54.7 & 32.1 & 19.8 & \textbf{71.0} \\
metal nut & 34.2 & 30.7 & 84.4 & \underline{87.8} & 87.1 & 77.0 & 86.0 & 65.2 & 64.6 & \textbf{94.0} \\
pill & 29.9 & 17.6 & 62.6 & 46.5 & \underline{79.5} & 63.7 & 73.5 & 22.8 & 45.5 & \textbf{90.8} \\
screw & 8.3 & 0.9 & \textbf{66.9} & \underline{63.8} & 37.8 & 58.7 & 47.2 & 14.8 & 0.7 & 50.9 \\
tile & 27.8 & 26.6 & \textbf{90.8} & \underline{88.5} & 77.8 & 71.8 & 69.4 & 69.9 & 61.4 & 86.2 \\
toothbursh & 25.1 & 4.7 & 47.5 & 37.2 & 62.1 & 58.7 & \underline{63.8} & 35.1 & 22.6 & \textbf{73.4} \\
transistor & 26.7 & 19.9 & 55.2 & 34.8 & \underline{76.3} & 59.8 & 64.2 & 28.9 & 33.2 & \textbf{85.7} \\
wood & 25.2 & 10.0 & \textbf{75.1} & 68.7 & 48.6 & 61.3 & 60.3 & 51.9 & 49.9 & \underline{74.5} \\
zipper & 26.8 & 4.5 & 68.2 & \underline{73.7} & 65.8 & 69.4 & 70.0 & 45.6 & 56.9 & \textbf{79.2} \\
average & 28.1 & 10.4 & \underline{66.7} & 62.3 & 61.4 & 66.2 & 65.2 & 42.8 & 43.4 & \textbf{76.4}\\
\bottomrule
\end{tabular}}
\caption{\textbf{Comparison on anomaly localization with $\mathbf{F_1}$-max.}}
\label{tab:comparison on anomaly localization with f_1 max. }
\end{table*}


\end{document}